\documentclass[sigconf]{acmart}
\AtBeginDocument{%
  \providecommand\BibTeX{{%
    \normalfont B\kern-0.5em{\scshape i\kern-0.25em b}\kern-0.8em\TeX}}}

\usepackage{graphicx}
\usepackage{amsmath}
\usepackage{amsfonts}
\usepackage{bm}
\usepackage{multirow}
\usepackage{enumitem}
\usepackage{arydshln}
\usepackage{booktabs}
\usepackage{multirow}
\usepackage{natbib}
\usepackage{graphicx}

\usepackage{pifont}
\usepackage{newunicodechar}
\usepackage{color}
\newunicodechar{✓}{\ding{51}}
\newunicodechar{✗}{\ding{55}}
\definecolor{greencheck}{RGB}{0,176,80}
\definecolor{redcross}{RGB}{255,0,0}

\DeclareMathOperator*{\argmax}{arg\,max}

\usepackage{xcolor}
\usepackage{bm}

\usepackage{algorithmic}
\usepackage[ruled,vlined,linesnumbered]{algorithm2e}
\let\oldnl\nl%
\newcommand{\nonl}{\renewcommand{\nl}{\let\nl\oldnl}}

\SetCommentSty{mycommfont}
\SetNlSty{emph}{}{}

\copyrightyear{2021} 
\acmYear{2021} 
\setcopyright{acmcopyright}\acmConference[SIGIR '21]{Proceedings of the 44th International ACM SIGIR Conference on Research and Development in Information Retrieval}{July 11--15, 2021}{Virtual Event, Canada}
\acmBooktitle{Proceedings of the 44th International ACM SIGIR Conference on Research and Development in Information Retrieval (SIGIR '21), July 11--15, 2021, Virtual Event, Canada}
\acmPrice{15.00}
\acmDOI{10.1145/3404835.3462853}
\acmISBN{978-1-4503-8037-9/21/07}

\begin{document}

\fancyhead{}

\title{Answering Any-hop Open-domain Questions \\ with Iterative Document Reranking}

\author{Ping Nie}
\authornote{Both authors contributed equally to this work.}
\affiliation{\institution{Peking University}\country{}}
\email{ping.nie@pku.edu.cn}

\author{Yuyu Zhang}
\authornotemark[1]
\affiliation{\institution{Georgia Institute of Technology}\country{}}
\email{yuyu@gatech.edu}

\author{Arun Ramamurthy}
\affiliation{\institution{Siemens Corporate Technology\country{}}}
\email{arun.ramamurthy@siemens.com}

\author{Le Song}
\affiliation{\institution{Georgia Institute of Technology\country{}}}
\email{lsong@cc.gatech.edu}

\begin{abstract}
Existing approaches for open-domain question answering (QA) are typically designed for questions that require either single-hop or multi-hop reasoning, which make strong assumptions of the complexity of questions to be answered. Also, multi-step document retrieval often incurs higher number of relevant but non-supporting documents, which dampens the downstream noise-sensitive reader module for answer extraction. To address these challenges, we propose a unified QA framework to answer any-hop open-domain questions, which iteratively retrieves, reranks and filters documents, and adaptively determines when to stop the retrieval process. To improve the retrieval accuracy, we propose a graph-based reranking model that perform multi-document interaction as the core of our iterative reranking framework. Our method consistently achieves performance comparable to or better than the state-of-the-art on both single-hop and multi-hop open-domain QA datasets, including Natural Questions Open, SQuAD Open, and HotpotQA.
\end{abstract}

\ccsdesc[500]{Information systems~Question answering}

\keywords{open-domain question answering; iterative document reranking; multi-document interaction}

\maketitle

\section{Introduction}

\setlength{\abovedisplayskip}{3pt}
\setlength{\abovedisplayshortskip}{3pt}
\setlength{\belowdisplayskip}{3pt}
\setlength{\belowdisplayshortskip}{3pt}
\setlength{\jot}{2pt}
\setlength{\floatsep}{1ex}
\setlength{\textfloatsep}{1ex}
\setlength{\intextsep}{1ex}

\begin{figure}[t]
\center
\includegraphics[width=0.48\textwidth]{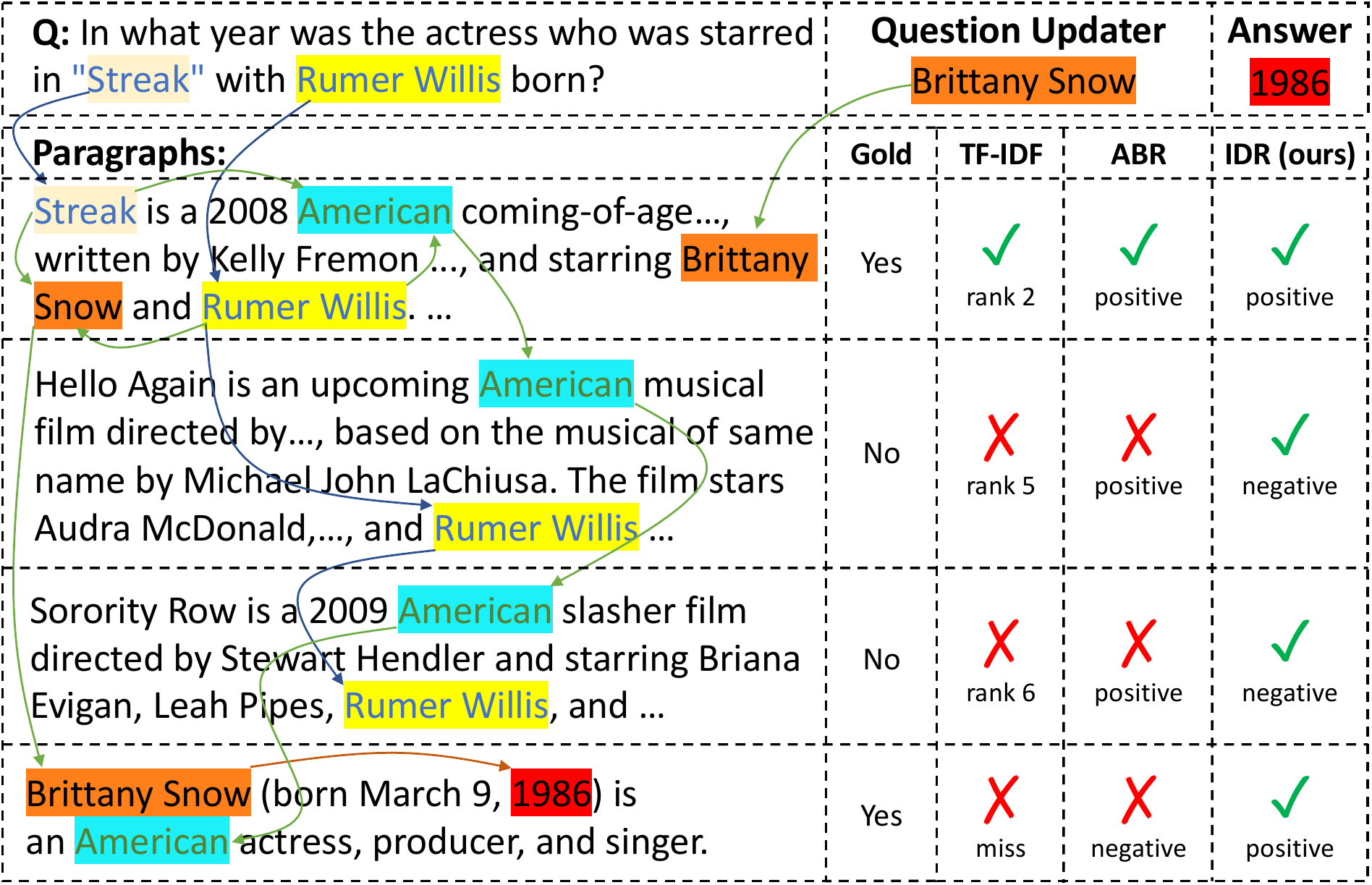}
\center
\caption{An example open-domain multi-hop question from the HotpotQA dev set, where the question has only partial clues to retrieve supporting documents. The first and fourth paragraphs are gold supporting documents, and the remaining two paragraphs are relevant but non-supporting documents. ABR stands for ALBERT-base reranker, which serves as a reference of the retrieval performance of existing multi-hop QA methods that independently consider the relevance of each document to the question. The \textbf{\color{greencheck}✓} and \textbf{\color{redcross}✗} symbols mark whether the retriever correctly identifies the document as a supporting / non-supporting one. Below each symbol, we annotate the output of the corresponding retriever with regard to the document, where ``positive'' (``negative'') means that the retriever classifies it as a supporting (non-supporting) document.}
\label{figure:example_of_qa_dev}
\end{figure}

Open-domain question answering (QA) requires a system to answer factoid questions using a large text corpus (e.g., Wikipedia or the Web) without any pre-defined knowledge schema. Most state-of-the-art approaches for open-domain QA follow the retrieve-and-read pipeline initiated by \citet{chen-etal-2017-reading}, using a retriever module to retrieve relevant documents, and then a reader module to extract answer from the retrieved documents. These approaches achieve prominent results on single-hop QA datasets such as SQuAD~\cite{rajpurkar-etal-2016-squad}, whose questions can be answered using a single supporting document. However, they are inherently limited to answering simple questions and not able to handle multi-hop questions, which require the system to retrieve and reason over evidence scattered among multiple documents. In the task of open-domain multi-hop QA~\cite{yang-etal-2018-hotpotqa}, the documents with the answer can have little lexical overlap with the question and thus are not directly retrievable. Take the question in Figure~\ref{figure:example_of_qa_dev} as an example, the last paragraph contains the correct answer but cannot be directly retrieved using TF-IDF. In this example, the single-hop TF-IDF retriever is not able to retrieve the last supporting paragraph since it has no lexical overlap with the question, but this paragraph contains the answer and plays a critical role in the reasoning chain.

Recent studies on multi-hop QA attempt to perform iterative retrievals to improve the answer recall of the retrieved documents. However, several challenges are not solved yet by existing multi-hop QA methods: 1) The iterative retrieval rapidly increases the total number of retrieved documents and introduces much noise to the downstream reader module for answer extraction. Typically, the downstream reader module is noise-sensitive, which works poorly when taking noisy documents as input or missing critical supporting documents with the answer~\cite{nie-etal-2019-revealing}. This requires the QA system to reduce relevant but non-supporting documents fed into the reader module. However, to answer open-domain multi-hop questions, it is necessary to iteratively retrieve documents to increase the overall recall of supporting documents. This dilemma poses a challenge for the retrieval phase of open-domain QA systems; 2) Existing multi-hop QA methods such as MUPPET~\cite{feldman-el-yaniv-2019-multi} and Multi-step Reasoner~\cite{das2018multistep} perform a fixed number of retrieval steps, which make strong assumptions on the complexity of open-domain questions and perform fixed number of retrieval steps. In real-world scenarios, open-domain questions may require different number of reasoning steps; 3) The relevance of each retrieved document to the question is independently considered. As exemplified in Figure~\ref{figure:example_of_qa_dev}, ABR stands for ALBERT-base reranker, which serves as a reference of the retrieval performance of existing multi-hop QA methods that independently consider the relevance of each document to the question. Without considering multiple retrieved documents as a whole, these methods can be easily biased to the lexical overlap between each document and the question, and incorrectly classify non-supporting documents as supporting evidence (such as the middle two non-supporting paragraphs in Figure~\ref{figure:example_of_qa_dev}, which have decent lexical overlap with the question) and vice versa (such as the bottom paragraph in Figure~\ref{figure:example_of_qa_dev}, which has no lexical overlap with the question but is a critical supporting document that contains the answer).

To address the challenges above, we introduce a unified QA framework for answering any-hop open-domain questions named Iterative Document Reranking (IDR). Our framework learns to iteratively retrieve documents with updated question, rerank and filter documents, and adaptively determine when to stop the retrieval process. In this way, our method can significantly reduce the noise introduced by multi-round retrievals and handle open-domain questions that require different number of reasoning steps. To avoid the bias of lexical overlap in identifying supporting documents, our framework considers the question and retrieved documents as a whole and models the multi-document interactions to improve the accuracy of classifying supporting documents.

\begin{figure*}
\center
\includegraphics[width=\textwidth]{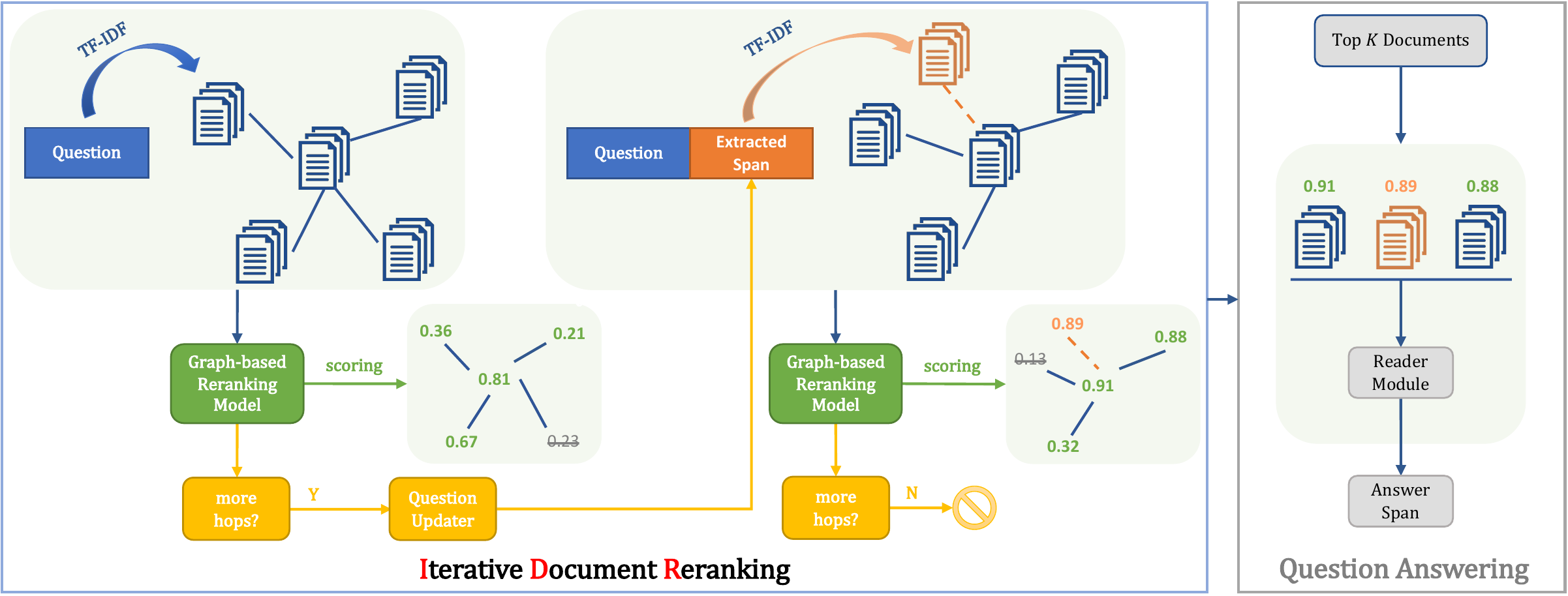}
\centering
\caption{An overview of the IDRQA system, which consists of an Iterative Document Reranking (IDR) phase and a question answering phase. Given an open-domain question, IDR iteratively retrieves, reranks and filters documents, and adaptively determines when to stop the retrieval process. After the initial retrieval, IDR updates the question with an extracted text span as a new query to retrieve more documents at every iteration. Once the retrieval is done, the final highest-scoring documents are fed into the downstream reader module for answer extraction.
}
\label{figure:pipeline}
\end{figure*}

As illustrated in Figure~\ref{figure:pipeline}, our method constructs a document graph linked by shared entities to propagate information using a graph attention network (GAT). By leveraging the multi-document information, our reranking model has more knowledge to differentiate supporting documents from irrelevant documents. After initial retrieval, our method updates the question at every retrieval step with a text span extracted from the retrieved documents, and then use the updated question as query to retrieve complementary documents, which are added to the document graph for a new round of interaction. The reranking model is reused to score the documents again and filter the most irrelevant ones. The maintained high-quality shortlist of remaining documents are then fed into the Reader Module to determine whether the answer exists in them. If so, the retrieval process ends and the QA system delivers the answer span extracted by the Reader Module as the predicted answer. Otherwise, the retrieval process continues to the next hop.

Our contributions are summarized as follows:
\begin{itemize}[leftmargin=*]
    \item \textit{Noise control for iterative retrieval}: We propose a novel QA method to iteratively retrieve, rerank and filter documents, and adaptively determine when to stop the retrieval process. Our method maintains a high-quality shortlist of remaining documents, which significantly reduces the noise introduced to the downstream reader module for answer extraction. Thus, the downstream reader module can extract the answer span with higher accuracy.
    \item \textit{Unified framework for any-hop open-domain QA}: We propose a unified framework that does not require to pre-determine the complexity of input questions. Different from existing QA methods that are specifically designed for either single-hop or fixed-hop questions, our method can adaptively determine the termination of retrieval and answer any-hop open-domain questions.
    \item \textit{Multi-document interaction}: We construct entity-linked document graph and employ graph attention network for multi-document interaction, which boosts up the reranking performance. To the best of our knowledge, we are the first to propose graph-based document reranking method for open-domain multi-hop QA.

\end{itemize}

\section{Overview}

\subsection{Problem Definition}
Given a factoid question, the task of open-domain question answering (QA) is to answer it using a large corpus which can have millions of documents (e.g., Wikipedia) or even billions (e.g., the Web). Let the corpus $\mathcal{C} = \{ d_1, d_2, \ldots, d_{|\mathcal{C}|} \}$ consist of $|\mathcal{C}|$ documents as the basic retrieval units\footnote{We use the natural paragraphs as the basic retrieval units.}. Each document $d_i$ can be viewed as a sequence of tokens $t_1^{(i)}, t_2^{(i)}, \ldots, t_{|d_i|}^{(i)}$. Formally, given a question $q$, the task is to find a text span $t_s^{(j)}, t_{s+1}^{(j)},\ldots, t_e^{(j)}$ from one of the documents $d_j$ that can answer the question\footnote{In this work, we focus on the extractive or span-based QA setting, but the problem definition and our proposed method can be generalized to other QA settings as well.}. For open-domain multi-hop QA, the final documents with the answer are typically multiple hops away from the question, i.e., the system is required to find seed documents and subsequent supporting documents in order of a chain or directed graph to locate the final documents. The retrieved documents are usually connected via shared entities or semantic similarities, and the formed chain or directed graph of documents can be viewed as the reasoning process for answering the question.

Note that the task of \textit{open-domain multi-hop QA} that we describe above is much different from the \textit{few-document setting of multi-hop QA}~\cite{qi-etal-2019-answering}, where the QA system is provided with a tiny set of documents that consists of all the gold supporting documents together with several irrelevant ``distractor'' documents. The \textit{few-document setting} is designed to test the system's capability of multi-hop reasoning given all of the gold supporting documents, but this is far from being realistic. A real-world open-domain QA system has to locate the necessary supporting documents from a large corpus on its own, which is especially challenging for multi-hop questions since the indirect supporting documents are not easily retrievable given the question itself.

The nature of multi-hop questions poses significant challenge for retrieving supporting documents, which is crucial to the downstream QA performance. To address this challenge, we argue that it is necessary to iteratively retrieve, rerank and filter documents, so that we can maintain a high-quality shortlist of documents. To this end, we propose the Iterative Document Reranking (IDR) method, which is introduced in the next section.

\subsection{System Overview}

\begin{algorithm}[t]
\caption{Iterative Document Reranking (inference)}
\label{alg:algorithm}
\DontPrintSemicolon
\SetNoFillComment
\SetKwInOut{Input}{input}\SetKwInOut{Output}{output}
\Input{A textual question $q$; the maximum number of retrieval hops $\mathcal{H}$; the number of retrieved documents at each hop $N$; the number of remaining documents after each reranking process $K$; the graph-based reranking model $\mathcal{M}$; the reader module $\mathcal{R}$; the question updater model $\mathcal{U}$}
\Output{Predicted answer $a$}
$h \gets 0$ \;
$\mathcal{D} \gets \text{\O}$ \;
\While{$h <= \mathcal{H}$}
{
    $h \gets h + 1$ \;
    $\mathcal{D}_{\text{new}} \gets$ Retrieve top $N$ documents according to $q$ \;
    $\mathcal{D} \gets \mathcal{D} \cup \mathcal{D}_{\text{new}}$ \;
    $\mathcal{D} \gets$ Rerank and get top $K$ documents using $\mathcal{M}(q, \mathcal{D})$ \;
    $a \gets$ Predict the answer using $\mathcal{R}(q, \mathcal{D})$ \;
    \If(\tcc*[f]{No more hop needed})
    {$a$ is not None}
    {
        \Return a \;
    }
    \Else(\tcc*[f]{More hop with question updater})
    {
        $q \gets \mathcal{U}(q, \mathcal{D})$ \;
    }
}
\Return a
\end{algorithm}

We illustrate the overview of our IDRQA system in Figure~\ref{figure:pipeline}. The IDRQA system first uses a given question as query to retrieve top $|\mathcal{D}|$ documents using TF-IDF. To construct the document graph,
IDR extracts the entities from the question and retrieved documents using an off-the-shelf Named Entity Recognition (NER) system and connects two documents if they have shared entities.
The graph-based reranking model takes the document graph as input to score each document, filter the lowest-scoring documents, and adaptively determines whether to continue the retrieval process. At every future retrieval step, IDR updates the question with an extracted text span from the retrieved documents as a new query to retrieve more documents.
Once the retrieval is done, the final highest-scoring documents are concatenated to feed into the downstream reader module \citep{devlin-etal-2019-bert,Lan2020ALBERT:} for answer extraction.

To describe the pipeline of our IDRQA system more precisely, we provide a concise algorithm that summarizes the inference procedure of the iterative document reranking (IDR) phase, as in Algorithm~\ref{alg:algorithm}. We maintain a set of retrieved documents $\mathcal{D}$. For each retrieval step, we retrieve the top $N$ documents according to the question $q$ and then append all the newly retrieved documents to the current set of retrieved documents. Then we use the graph-based reranking model $\mathcal{M}$ to rerank those documents, get top $K$ of them, and filter the other ones. This shortlist of retrieved documents together with the question $q$ are then fed in to the downstream reader module $\mathcal{R}$ for answer extraction. Note that the reader module has the option to predict that there is no answer to be extracted from the provided documents. If the reader module does predict no answer, we use the question updater model $\mathcal{U}$ to update the question with an extracted span and move to the next hop of retrieval. The while loop continues until a valid answer is extracted by the reader module or the maximum number of retrieval hops $\mathcal{H}$ is reached.

\begin{figure*}[ht]
\center
\includegraphics[width=\textwidth]{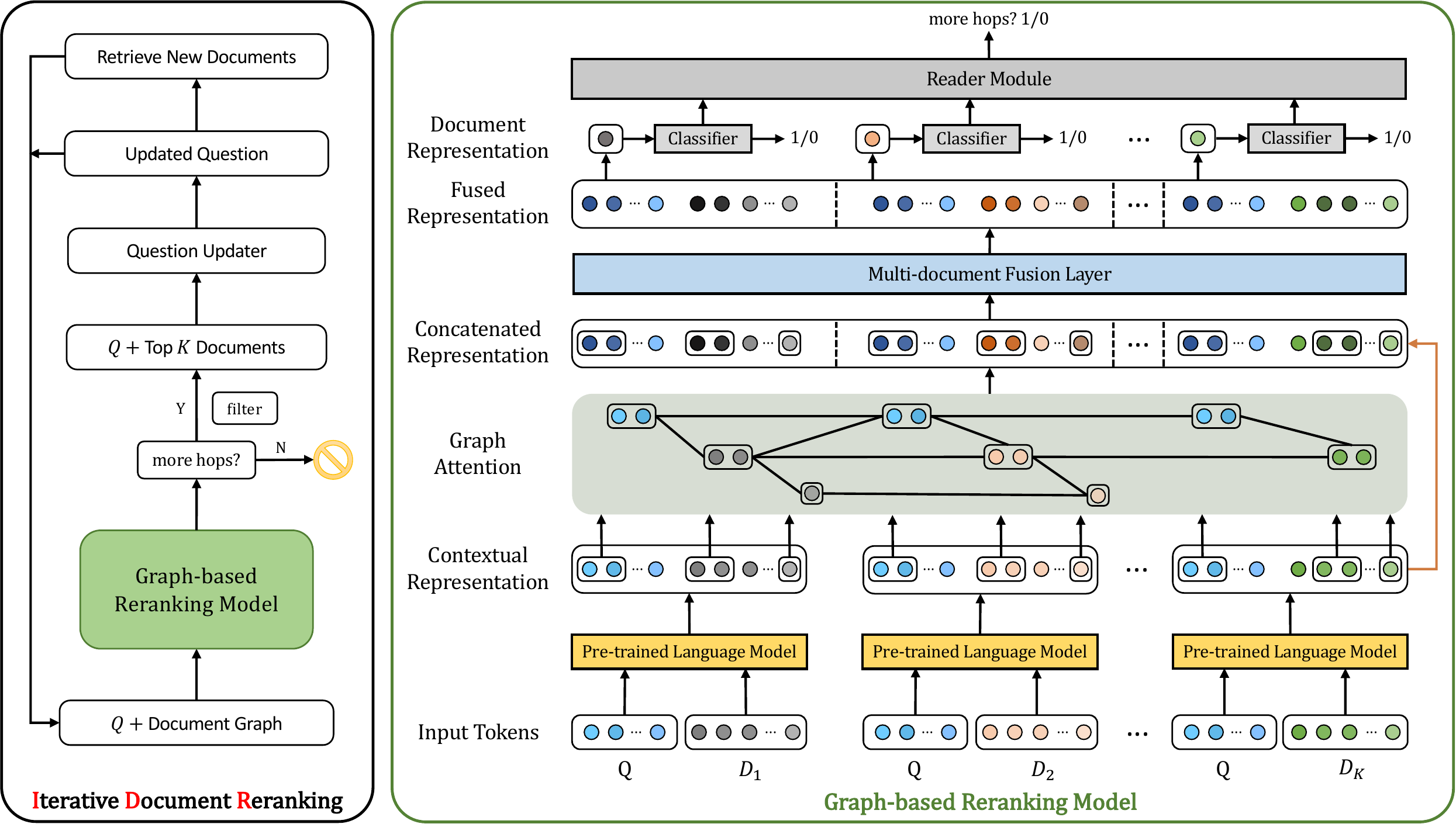}
\center
\caption{As the core of our IDR framework, the Graph-based Reranking Model first encodes the question and each retrieved document with pre-trained language model to generate contextual representations, and uses the shared entities to propagate information using a Graph Attention Network (GAT). After the entity-entity interaction across multiple documents, the updated entity representations with the original contextual encodings are fed into the fusion layer for further interaction. Finally, the reranking model takes pooled document representations to score each document and filter low-scoring documents. The maintained high-quality shortlist of remaining documents are then fed into the Reader Module to determine whether the answer exists in them. If so, the retrieval process ends and the QA system delivers the answer span extracted by the Reader Module as the predicted answer. Otherwise, the retrieval process continues to the next hop.}
\label{figure:IDR}
\end{figure*}

\section{IDRQA System} \label{sec:ddr}

\subsection{Graph-based Reranking Model} \label{subsec:graph-ranking-model}

The graph-based reranking model (Figure~\ref{figure:IDR}) is designed to precisely identify the supporting documents in the document graph. We present the components of this reranking model as follows.

\BlankLine
\noindent \textbf{Contextual Encoding.}
Given a question $q$ and $|\mathcal{D}|$ documents retrieved by TF-IDF, we concatenate the tokens of the question and each document to feed into the pre-trained language model as:
\begin{equation}
	\mathcal{I}_{q,d_k} = \mathrm{[CLS]} \ q_1 \ldots q_{|q|} \ \mathrm{[SEP]} \ t_1^{(k)} \ldots t_{|d_k|}^{(k)} \  \mathrm{[SEP]}, \nonumber
\end{equation}
where $|q|$ and $|d_k|$ denote the number of tokens in the question $q$ and the document $d_k$, respectively. $\mathrm{[CLS]}$ and $\mathrm{[SEP]}$ are special tokens used in pre-trained language models such as BERT~\cite{devlin-etal-2019-bert} and ALBERT~\cite{Lan2020ALBERT:}. Thus we independently encode each document $d_k$ along with the question $q$ to obtain the contextual representation vector $\mathbf{v}_{q,d_k} \in \mathbb{R}^{L \times h}$, where $L$ is the maximum length of the input tokens $\mathcal{I}$, and $h$ is the embedding size.
For efficient batch computation, we pad or truncate the input tokens to the length of $L$.
We then concatenate all documents' contextual representation vectors as $\mathbf{v} \in \mathbb{R}^{L |\mathcal{D}| \times h}$.

\BlankLine
\noindent \textbf{Graph Attention.}
After we obtain the question-dependent encoding of each document, we employ a Graph Attention Network (GAT; \citet{velickovic2018graph}) to propagate information on the document graph, where two documents are connected if they have shared entities. To be more specific, for each shared entity $E_i$ in the document graph, we perform pooling over its token embeddings from $\mathbf{v}$ to produce the entity embedding as $\mathbf{e}_i = \mathrm{Pooling} \big( \mathbf{t}_1^{(i)}, \mathbf{t}_2^{(i)}, \ldots, \mathbf{t}_{|E_i|}^{(i)} \big) $ where $\mathbf{t}_j^{(i)}$ is the embedding of the $j$-th token in $E_i$, and $|E_i|$ is the number of tokens in $E_i$. We use both mean- and max-pooling, thus we have $\mathbf{e}_i \in \mathbb{R}^{2h}$. Inspired by \citet{qiu-etal-2019-dynamically}, we apply a dynamic soft mask on the entities, serving as the information ``gatekeeper'' which assigns more weights to entities pertaining to the question. The soft mask applied on each entity $E_i$ is computed as
\begin{align}
	m_i &= \sigma \Bigg( \frac{\mathbf{q} V \mathbf{e}_i}{\sqrt{2h}} \Bigg) \\
	\mathbf{g}_i &= m_i \mathbf{e}_i,
\end{align}
where $\mathbf{q} \in \mathbb{R}^{2h}$ is the concatenated mean- and max-pooling of the question token embeddings, and $V \in \mathbb{R}^{2h \times 2h}$ is a linear projection matrix, $\sigma(\cdot)$ is the sigmoid function, and $\mathbf{g}_i$ is the masked entity embedding.
We then use GAT to disseminate information between entities. Starting from $\mathbf{g}_i^{(0)} = \mathbf{g}_i$, GAT iteratively updates the embedding of each entity with the information from its neighbors as
\begin{align}
    \mathbf{h}_i^{(t)} &= W_i \mathbf{g}_i^{(t-1)} + b_i \\
    \mathbf{g}_i^{(t)} &= \mathrm{ReLU}(\sum_{j\in N(i)} \alpha_{i, j}^{(t)} \mathbf{h}_j^{(t)}),
\end{align}
where $\mathbf{h}_i^{(t)}$ denotes the hidden states of $E_i$ on the $t$-th GAT layer, $W_1 \in \mathbb{R}^{h \times 2h}$ is a linear projection matrix, $b_i$ is a bias term, $N(i)$ is the set of neighbor entities of $E_i$, and the entity-entity attention $\alpha_{i, j}^{(t)}$ is computed as follows:
\begin{align}
    s_{i, j}^{(t)} &= \mathrm{LeakyReLU}(W_{2} [\mathbf{h}_i^{(t)} ;\mathbf{h}_j^{(t)}]) \\
    \alpha_{i,j}^{(t)} &= \frac{\exp(s_{i,j}^{(t)})}{\sum_k \exp(s_{i,k}^{(t)})},
\end{align}
where $W_{2} \in \mathbb{R}^{2h}$ is a linear projection matrix. 
We finally obtain the GAT updated entity embeddings $\mathbf{g}_i^{(T)}$, where $T$ is the number of GAT layers.

\BlankLine
\noindent \textbf{Multi-document Fusion.}
To further propagate the information to non-entity tokens, we firs use the embedding of each entity token as
\begin{equation}
    \hat{\mathbf{t}}_j^{(i)} =  W_3 \big[ \mathbf{t}_j^{(i)} ; \mathbf{g}_i^{(T)} \big],
\end{equation}
where $W_{3} \in \mathbb{R}^{h \times 3h}$ is a linear projection matrix. Then we replace the corresponding vectors in $\mathbf{v}$ with $\hat{\mathbf{t}}_j^{(i)}$ to obtain $\hat{\mathbf{v}} \in \mathbb{R}^{L |\mathcal{D}| \times h}$. Finally, $\hat{\mathbf{v}}$ is fed into a Transformer \citep{vaswani2017attention} layer for multi-document fusion, which updates the representations of all the tokens and outputs the fused representation vectors $\tilde{\mathbf{v}} \in \mathbb{R}^{L |\mathcal{D}| \times h}$.

\BlankLine
\noindent \textbf{Document Filter.}
For each document, we use the $\mathrm{[CLS]}$ token embedding from $\tilde{\mathbf{v}}$ as the document representation, which is fed into a binary classifier to score the document's supporting level. In each retrieval hop, the top $K$ documents with the highest scores are selected and the rest are filtered.

\subsection{Question Updater}

In open-domain multi-hop QA, the question seldom contains all the retrievable clues and one has to identify the missing information to proceed with further reasoning~\cite{yang-etal-2018-hotpotqa}. To increase the recall of indirect supporting documents, we integrate the query generator of GoldEn Retriever~\citep{qi-etal-2019-answering} into our system, which serves as the question updater at every retrieval step after the initial one.

Question Updater comes after Reader module if no answer was found in top $K$ documents. It aims to generate the \textit{clue span} other than the answer span from the reranked top $K$ documents given current question $q$. Thus conceptually, this process is similar to the Reader Module which extracts answer span from retrieved documents. More specifically, following GoldEn Retriever~\citep{qi-etal-2019-answering}, we use the same method for data construction to train a span extractor model, which extracts the \textit{clue span} from retrieved documents by predicting its start and end tokens. Formally, given the question $q$ and the reranked top $K$ documents $\mathcal{D}_K = \{d_1, d_2, \ldots, d_K\}$, the Question Updater generates the \textit{clue span} $C$ and concatenate $C$ with the original question $q$ as
\begin{align}
    C &=\mathcal{U}(q, \mathcal{D}_K) \\
    q &= \text{concatenate}(q, \mathrm{[SEP]}, C),
\end{align}
where $\mathrm{[SEP]}$ is a special separator token. In order to train the Question Updater, we construct each training sample as a triple of $<q, \mathcal{D}_K, C>$, where $q$ is the question, $\mathcal{D}_K=\{d_1, d_2, \ldots, d_K\}$ is the set of top $K$ retrieved documents reranked by our graph-based reranking model.

\subsection{Reader Module}
In this work, we mainly focus on the retrieval phase, and use a standard span-based reader module as in BERT~\citep{devlin-etal-2019-bert} and ALBERT~\citep{Lan2020ALBERT:}. We concatenate the tokens of the question $q$ and the final top $K$ reranked documents to feed into the reader module. At inference time, the reader finds the best candidate answer span by
\begin{equation}
    \argmax_{i, j, \ i \leq j} P_i^{\mathrm{start}} P_j^{\mathrm{end}},
\end{equation}
where $P_i^{\mathrm{start}}$, $P_j^{\mathrm{end}}$ denote the probability that the $i$-th and $j$-th tokens are the start and end positions in the concatenated text, respectively, of the answer span. During inference, there is no guarantee that the answer span exists in the reader's input text. To handle the no-answer cases, the reader predicts $P_{na}$ as the probability of having no answer span, and compares $P_{na}$ with $\max_{i, j, \ i \leq j} P_i^{\mathrm{start}} P_j^{\mathrm{end}}$ to determine the output between a special no-answer prediction and the best candidate answer span.

\section{Experiments}

\subsection{Experimental Settings} \label{graph-base reranking setting}
We conduct all the experiments on a GPU-enabled (Nvidia RTX 6000) Linux machine powered by Intel Xeon Gold 5125 CPU at 2.50GHz with 384 GB RAM. For the graph-based reranking model, we set the maximum token sequence length $L = 250$, the number of retrieved documents $|\mathcal{D}| = 8$, the maximum number of entities $|\mathcal{E}| = 120$. The embedding size $h$ is $768$ and $4,096$ for the ALBERT-base and ALBERT-xxlarge model, respectively. The graph attention module has $T=2$ GAT layers. The maximum number of retrieval hops $\mathcal{H}=4$. The top $K=4$ reranked paragraphs are sent into the downstream reader module.

We implement our system based on HuggingFace's Transformers~\citep{Wolf2019HuggingFacesTS}. Following the previous state-of-the-art method~\cite{Fang2019HierarchicalGN}, we use ALBERT-xxlarge~\citep{Lan2020ALBERT:} as the pre-trained language model. We use AdamW \citep{Wolf2019HuggingFacesTS} as the optimizer and tune the initial learning rate between $1.5\mathrm{e}{-5}$ and $2.5\mathrm{e}{-5}$.

\subsection{Datasets} \label{appdx:data}

\setlength\tabcolsep{16pt}
\begin{table*}[]
\centering
\caption{Performance on the HotpotQA full wiki dev set. }
\label{table:overall qa performance on dev}
\begin{tabular}{@{}lccccc@{}}
\toprule
\multirow{2}{*}{Method} & \multicolumn{2}{c}{Answer} & \multicolumn{2}{c}{Supporting Fact} & Paragraph \\ \cmidrule(l){2-6} 
 & EM & F$_1$ & EM & F$_1$ & EM \\ \midrule
Cognitive Graph \citep{ding-etal-2019-cognitive} & 37.6 & 49.4 & 23.1 & 58.5 & 57.8 \\
DecompRC \citep{min-etal-2019-multi} & -- & 43.3 & -- & -- & -- \\
MUPPET \citep{feldman-el-yaniv-2019-multi} & 31.1 & 40.4 & 17.0 & 47.7 &  \\
GoldEn Retriever \citep{qi-etal-2019-answering} & -- & 49.7 & -- & -- & -- \\
DrKIT \citep{Dhingra2020Differentiable} & 35.7 & 46.6 & -- & -- & -- \\
Semantic Retrieval MRS \citep{nie-etal-2019-revealing} & 46.5 & 58.8 & 39.9 & 71.5 & 63.9 \\
Transformer-XH \citep{zhaotransxh2020} & 50.2 & 62.4 & 42.2 & 71.6 & -- \\
Recurrent Retriever \citep{asai2020learning} & 60.5 & 73.3 & 49.3 & 76.1 & 75.7 \\
HopRetriever \citep{li2020hopretriever} & 62.1 & 75.2 & \textbf{52.5} & 78.9 & \textbf{82.5} \\
\midrule
IDRQA (ours) & \textbf{62.9} & \textbf{75.9} & 51.3 & \textbf{79.1} & 79.8 \\ \bottomrule
\end{tabular}
\end{table*}

\setlength\tabcolsep{14pt}
\begin{table*}[]
\centering
\caption{Performance on the HotpotQA full wiki test set.}
\label{table:overall QA performance on test}
\begin{tabular}{@{}lcccccc@{}}
\toprule
\multirow{2}{*}{Method} & \multicolumn{2}{c}{Answer} & \multicolumn{2}{c}{Supporting Fact} & \multicolumn{2}{c}{Joint} \\ \cmidrule(l){2-7} 
 & EM & F$_1$ & EM & F$_1$ & EM & F$_1$ \\ \midrule
DecompRC \citep{min-etal-2019-multi} & 30.0 & 40.6 & -- & -- & -- & -- \\
QFE \citep{nishida-etal-2019-answering} & 28.6 & 38.0 & 14.2 & 44.3 & 8.6 & 23.1 \\
Cognitive Graph \citep{ding-etal-2019-cognitive} & 37.1 & 48.8 & 22.8 & 57.6 & 12.4 & 34.9 \\
MUPPET \citep{feldman-el-yaniv-2019-multi} & 30.6 & 40.2 & 16.6 & 47.3 & 10.8 & 27.0 \\
GoldEn Retriever \citep{qi-etal-2019-answering} & 37.9 & 48.5 & 30.6 & 64.2 & 18.0 & 39.1 \\
DrKIT \citep{Dhingra2020Differentiable} & 42.1 & 51.7 & 37.0 & 59.8 & 24.6 & 42.8 \\
Semantic Retrieval MRS \citep{nie-etal-2019-revealing} & 45.3 & 57.3 & 38.6 & 70.8 & 25.1 & 47.6 \\
Transformer-XH \citep{zhaotransxh2020} & 51.6 & 64.0 & 40.9 & 71.4 & 26.1 & 51.2 \\
Recurrent Retriever \citep{asai2020learning} & 60.0 & 73.0 & 49.0 & 76.4 & 35.3 & 61.1 \\
Hierarchical Graph Network \citep{Fang2019HierarchicalGN} & 59.7 & 71.4 & 51.0 & 77.4 & 37.9 & 62.2 \\
HopRetriever \citep{li2020hopretriever} & 60.8 & 73.9 & 53.1 & 79.3 & 38.0 & 63.9 \\
Multi-hop Dense Retrieval \citep{xiong2021answering} & 62.3 & 75.3 & \textbf{57.5} & \textbf{80.9} & \textbf{41.8} & \textbf{66.6} \\
\midrule
IDRQA (ours) & \textbf{62.5} & \textbf{75.9} & 51.0 & 78.9 & 36.0 & 63.9 \\ \bottomrule
\end{tabular}
\end{table*}

\setlength\tabcolsep{14pt}
\begin{table}[ht]
\centering
\caption{Answer EM scores on the test set of Natural Questions Open and SQuAD Open.}
\label{table:single hop QA performance on test}
\begin{tabular}{@{}lcc@{}}
\toprule
Method & NQ & SQuAD \\ \midrule
DrQA \citep{chen-etal-2017-reading} & -- & 29.8 \\
R$^3$ \citep{wang2018r} & -- & 29.1 \\
Multi-step Reasoner \citep{das2018multistep} & -- & 31.9 \\
BERTserini \citep{yang-etal-2019-end-end-open} & -- & 38.6 \\
MUPPET \citep{feldman-el-yaniv-2019-multi} & -- & 39.3 \\
Multi-passage BERT \citep{wang-etal-2019-multi} & -- & 53.0 \\
ORQA \citep{lee-etal-2019-latent} & 33.3 & 20.2 \\
Recurrent Retriever \citep{asai2020learning} & 31.7 & \textbf{56.5} \\
Dense Passage Retriever \citep{karpukhin2020dense} & 41.5 & 36.7 \\
\midrule
IDRQA (ours) & \textbf{45.5} & \textbf{56.6}  \\ \bottomrule
\end{tabular}                                                                  
\end{table}

We evaluate our method on three open-domain QA benchmark datasets: HotpotQA~\cite{yang-etal-2018-hotpotqa}, Natural Questions Open~\cite{lee-etal-2019-latent} and SQuAD Open~\cite{chen-etal-2017-reading}. On all the three datasets, we focus on the \textit{full wiki} open-domain QA setting, which requires the system to retrieve evidence paragraphs from the entire Wikipedia and extract the answer span from the retrieved paragraphs.

Following the train / dev / test splits of Natural Questions Open and SQuAD Open in previous works~\cite{lee-etal-2019-latent,karpukhin2020dense}, we use the original validation set as our test set and keep 10\% training set as our dev set. Natural Questions Open and SQuAD open consist of single-hop questions, while HotpotQA consists of 113K crowd-sourced multi-hop questions that require Wikipedia introduction paragraphs to answer.
In the train and dev splits of HotpotQA, each question for training comes with two gold supporting paragraphs annotated by the crowd workers. Thus we can evaluate the retrieval performance on the dev set of HotpotQA dataset.

\subsection{Data Preprocessing}

Here we describe the details of data preprocessing methods that we develop for the HotpotQA dataset.

\BlankLine
\noindent \textbf{Training data construction for the graph-based reranking model.}
Graph-based reranking model aims to precisely score the retrieved documents by considering multiple documents as a whole instead of independent instances. In order to make our model reusable and robust to new test cases, we carefully design the training data construction method. To keep the distribution of training and test data consistent, we add negative samples to the training data for our graph-based reranking model. Formally, We pair each question $q$ with a set of documents $\mathcal{D}_{\mathrm{train}}$ to form a training sample. We design each training sample with the following strategies: 1) For each training sample which contains all supporting documents, we pair the question $q$ with $\mathcal{D}_{\mathrm{train}}$ where $\mathcal{D}_{\mathrm{train}}$ includes all $NP_s$ supporting documents necessary for multi-hop QA and $|\mathcal{D}_{\mathrm{train}}| - NP_s$ noisy documents. $\mathcal{D}_{\mathrm{train}}$ is a set of documents in training. $NN_s, NP_s$ are the number of noisy documents and the number of supporting documents; 2) For each training sample which contains partial supporting documents, we paired question $q$ with $\mathcal{D}_{\mathrm{train}}$ where $\mathcal{D}_{\mathrm{train}}$ contains $NP_s$ supporting documents and $|\mathcal{D}_{\mathrm{train}}| - NP_s$ noisy documents.$NP_s$ supporting documents are randomly sampled from all supporting documents; 3) For noisy documents, we sample them according to their score given by TF-IDF; 4) For multi-step reranking, we also randomly sample 30\% questions and use Question Updater to generate \textit{the clue span} and update questions. These new questions and their original paired documents are also used as training samples; 5) We concatenate all documents in $\mathcal{D}_{\mathrm{train}}$ in random order, rather than the reranked order. We use $|\mathcal{D}_{\mathrm{train}} | = 6$ and $NP_s = 2$ in our experiments.

\BlankLine
\noindent \textbf{Training data construction for the reader module.}
We designed the training data for the Reader Module carefully since the Reader module is not only used to predict the final answer but also to tell that \textit{no answer} found in in current context. The training sample is a triple of $<q, \mathcal{D}, a>$ where $q$ is question, $\mathcal{D} = \{d_1, d_2, d_3, d_4\}$ is 4 documents feed into QA model and $a$ is the final answer. For each question, we construct 5 types of training sample: 1) We concatenate two supporting passages (ordered as originally in the dataset) and two highest TF-IDF scored negative passages (ordered from higher to lower reranking score) as training samples; 2) We construct a random shuffled version of the 1st type in passage level; 3) We randomly replace one of the supporting passages of the 1st type by a negative passage; 4) One passage which has the final answer and is not one of the supporting passage and three negative passages; 5) We construct four negative passages with high TF-IDF score which are not supporting passages and do not contain the final answer.

\subsection{Evaluation Metrics}

We evaluate both the paragraph reranking performance and the overall QA performance. Following the existing studies~\cite{asai2020learning,nie-etal-2019-revealing,nishida-etal-2019-answering}, we evaluate the paragraph-level retrieval accuracy using the Paragraph Exact Match (EM) metric, which compares the top 2 paragraphs with the gold supporting paragraphs. For the QA performance, we report standard answer Exact Match (EM) and F$_1$ scores to measure the overlap between the gold answer and the extracted answer span.

\subsection{Overall Results}

We evaluate our method on both single-hop and multi-hop open-domain QA datasets. Note that our method is hop-agnostic, i.e., we consistently use the same setting of $\mathcal{H}=4$ as the maximum number of retrieval hops for all the three datasets.

For the multi-hop QA task, we report performance on both the dev and test sets of the HotpotQA dataset.
As reported in Table~\ref{table:overall qa performance on dev}, we compare the performance of our system IDRQA with various existing methods on the HotpotQA full wiki dev set. Since the golden supporting paragraphs are only labeled on the train / dev splits of the HotpotQA dataset, we can only report the paragraph EM metric on the dev set.
In Table~\ref{table:overall QA performance on test}, we compare our system with various existing methods on HotpotQA full wiki test set. IDRQA outperforms all published and previous unpublished methods on the HotpotQA dev set and the hidden test set on the official leaderboard (on May 21, 2020). 
Note that we focus on the QA task in this paper, and for the supporting fact prediction task we simply concatenate each pair of question and supporting fact to train a binary classification model. The joint EM and F$_1$ scores combine the evaluation of answer spans and supporting facts as detailed in \cite{yang-etal-2018-hotpotqa}. Thus we are behind state-of-the-art performance on supporting fact and joint scores.

For the single-hop QA task, we evaluate our method on Natural Questions (NQ) Open and SQuAD Open datasets. We summarize the performance in Table~\ref{table:single hop QA performance on test}. For NQ Open, we follow the previous work DPR~\cite{karpukhin2020dense} to use dense retriever instead of TF-IDF for document retrieval. Our method also achieves QA performance comparable to or better than the state-of-the-art methods on both datasets, which shows the robustness of our method across different open-domain QA datasets.

\begin{figure*}[ht]
\center
\includegraphics[width=.95\textwidth]{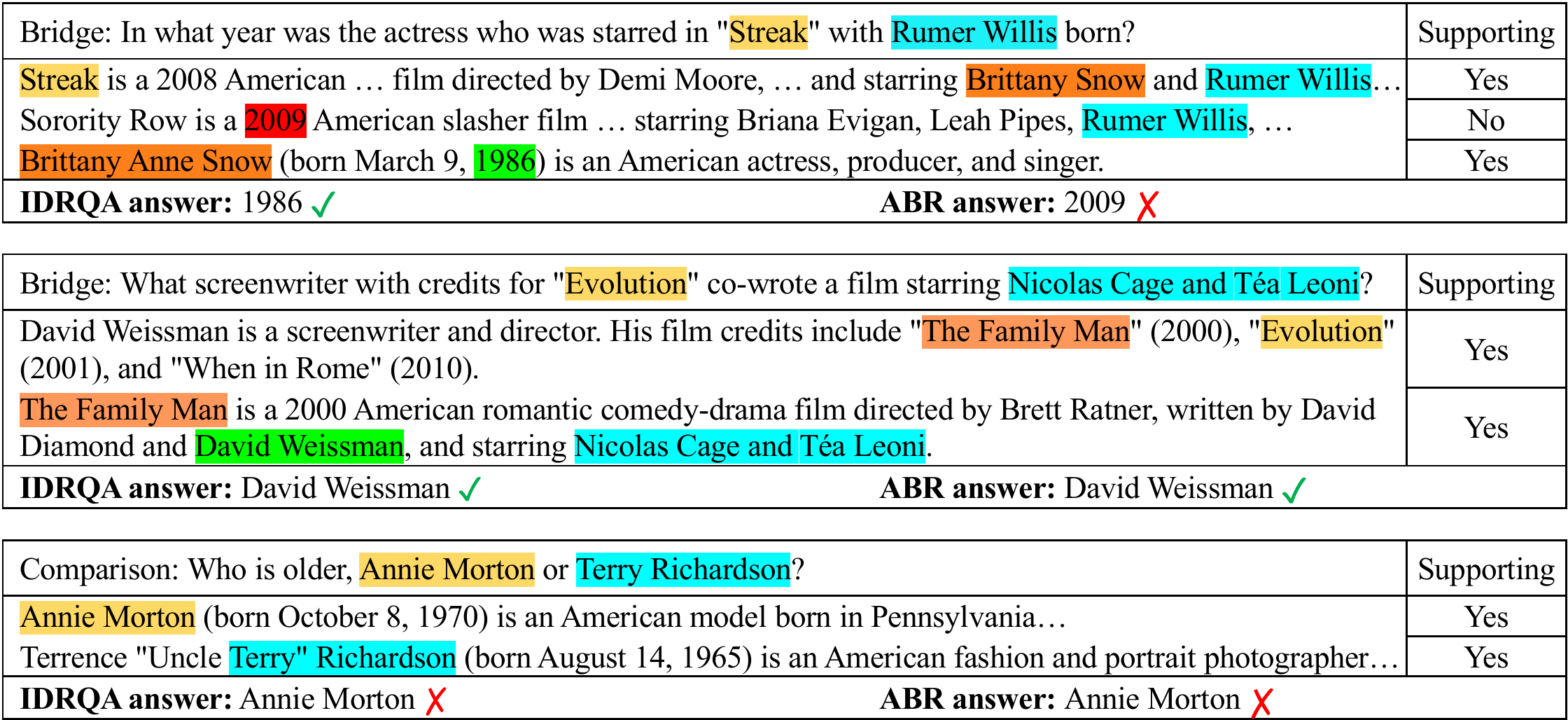}
\center
\caption{Case study of example questions with supporting paragraphs from HotpotQA dev set.}
\label{figure:case study}
\end{figure*}

\setlength\tabcolsep{12pt}
\begin{table*}
\centering
\caption{Ablation study of our system in different settings on HotpotQA full wiki dev set.}
\label{table:ablation comparison and bridge}
\begin{tabular}{@{}lccccccc@{}}
\toprule
\multirow{3}{*}{Ablation Setting} & \multicolumn{2}{c}{Bridge (79.9\%)} & \multicolumn{2}{c}{Comparison (20.1\%)} & \multicolumn{3}{c}{Full Dev (100\%)} \\ \cmidrule(l){2-8} 
 & \multicolumn{2}{c}{Answer} & \multicolumn{2}{c}{Answer} & \multicolumn{2}{c}{Answer} & Paragraph \\ \cmidrule(l){2-8} 
 & EM & F$_1$ & EM & F$_1$ & EM & F$_1$ & EM \\ \midrule
IDRQA & 58.1 & 73.2 & 71.4 & 77.9 & 60.7 & 74.2 & 73.8 \\
\ \ \ w/o Graph-based Reranking & 47.4 & 59.5 & 70.1 & 76.4 & 52.1 & 62.9 & 62.6 \\
\ \ \ w/o Iterative Reranking & 49.8 & 61.7 & 70.5 & 77.3 & 54.3 & 64.5 & 65.2 \\
\ \ \ w/o Question Updater & 56.6 & 71.4 & 71.3 & 77.8 & 59.8 & 72.9 & 70.3 \\ \bottomrule
\end{tabular}
\end{table*}

\subsection{Detailed Analysis}

\BlankLine
\noindent \textbf{Ablation study.}
To investigate the effectiveness of each module in IDRQA, we compare the performance of several variants of our system on HotpotQA full wiki dev set.
As shown in Table \ref{table:ablation comparison and bridge}, once we disable the \emph{Iterative Reranking}, \emph{Graph-based Reranking} and \emph{Question Updater} module, both the paragraph reranking and QA performance drop significantly. Notably, there is a 11 points drop in Paragraph EM decrease and a 12 points drop in Answer F$_1$ when \emph{Graph-based Reranking} is removed from IDRQA. This shows the importance of the graph-based reranking model in our system.
To further study the impact of these modules, we decompose the QA performance into question categories \textit{Bridge} and \textit{Comparison}. We find that the QA performance on the bridge questions drops much more significantly than that on the comparison questions. This is because the comparison questions require to compare two entities mentioned in the question~\cite{yang-etal-2018-hotpotqa}, thus iterative retrieval and multi-hop reasoning may not be necessary. In contrast, to answer the bridge questions which often have missing entities, our iterative graph-based document reranking method is of crucial importance.

\BlankLine
\noindent \textbf{Impact of retrieval steps.}
IDRQA aggregates the document scores to check whether the collected evidence is enough to answer the question, and adaptively determines when to stop the retrieval process. We investigate the number of retrieval steps selected by IDRQA, and report its distribution with breakdown performance in Table~\ref{table:stats_dynamic}. Over 60\% questions are answered with 2-step retrieval. About 20\% questions are answered with 1-step retrieval, which is close to the ratio of the comparison questions that may not need iterative retrieval. For questions that IDRQA selects to perform over 2-step retrieval, a significant drop on both reranking and QA performance is observed, showing that these questions are the hardest ones in HotpotQA.

\setlength\tabcolsep{6pt}
\begin{table}
\centering
\caption{Distribution of the selected retrieval steps on HotpotQA dev set, which is adaptively determined by IDRQA.}
\label{table:stats_dynamic}
\begin{tabular}{@{}lcccc@{}}
\toprule
\multirow{2}{*}{Retrieval} & \multirow{2}{*}{\% of Questions} & Paragraph & \multicolumn{2}{c}{Answer} \\ \cmidrule(l){3-5} 
 &  & EM & EM & F$_1$ \\ \midrule
1-step & 21.8 & 88.4 & 70.3 & 81.9 \\
2-step & 63.2 & 76.9 & 62.0 & 75.9 \\
3-step & 5.3 & 39.6 & 45.2 & 59.8 \\
4-step (max) & 9.7 & 22.0 & 37.7 & 50.4 \\ \bottomrule
\end{tabular}
\end{table}

\BlankLine
\noindent \textbf{Case study and limitations.}
We showcase several example questions with answers from IDRQA and the baseline ALBERT-base reranker (ABR) in Figure~\ref{figure:case study}. The first case is a hard bridge question where ABR extracts the wrong answer from a relevant but non-supporting paragraph, showing the advantage of iterative reranking in our system.
The second question is correctly answered by both IDRQA and ABR, since it provides sufficient clues to retrieve both paragraphs.
The final case is a comparison question that requires numerical reasoning, which is not correctly answered by both IDRQA and ABR. This shows the limitation of our system, and we plan to explore the combination of multi-hop and numerical reasoning in future work.

\section{Related Work}

\noindent \textbf{Open-domain QA.}  For text-based QA, the QA system is provided with semi-structured or unstructured text corpora as the knowledge source to find the answer. Text-based QA dates back to the QA track evaluations of the Text REtrieval Conference (TREC)~\cite{voorhees2000building}. Traditional approaches for text-based QA typically use a pipeline of question parsing, answer type classification, document retrieval, answer candidate generation, and answer reranking, such as the famous IBM Watson system which beat human players on the \textit{Jeopardy!} quiz show~\cite{ferrucci2010building}. However, such pipeline-based QA systems require heavy engineering efforts and only work on specific domains.
The open-domain QA task was originally proposed and formalized in \citet{chen-etal-2017-reading}, which builds a simple pipeline with a TF-IDF retriever module and a RNN-based reader module to produce answers from the top 5 retrieved documents. Different from the machine reading comprehension task (MRC) that provides a single paragraph or document as the evidence~\cite{rajpurkar-etal-2016-squad}, open-domain QA is more challenging since the retrieved documents are inevitably noisy. Recent works on open-domain QA largely follow the retrieve-and-read approach, and have made prominent improvement on both the retriever module~\cite{lee-etal-2019-latent,wang2018r,nie-etal-2019-revealing} and the reader module~\cite{wang2018evidence,wang2019sigir,ni-etal-2019-learning}. These approaches simply perform one-shot document retrieval to handle single-hop questions. However, for complicated questions that require multi-hop reasoning, these approaches are not applicable since they fail to collect necessary evidence scattered among multiple documents.

\BlankLine
\noindent \textbf{Open-domain multi-hop QA.} HotpotQA~\cite{yang-etal-2018-hotpotqa} is crowd-sourced over Wikipedia as the largest free-form dataset for open-domain multi-hop QA to date. Recently, a variety of approaches have been proposed to address the multi-hop challenge.
DecompRC~\cite{min-etal-2019-multi} decomposes a multi-hop question into simpler sub-questions and leverages single-hop QA models to answer it, which still uses one-shot TF-IDF retrieval to collect relevant documents. BERT Reranker~\cite{das-etal-2019-multi}, DrKIT~\cite{Dhingra2020Differentiable} and Transformer-XH~\cite{zhaotransxh2020} employ the entities in the question and the retrieved documents to link additional documents, which expands the one-shot retrieval results and improves the evidence coverage. GoldEn Retriever~\cite{qi-etal-2019-answering} adopts iterative TF-IDF retrieval by generating a new query for each retrieval step. Graph Retriever~\cite{min2019knowledge} employs entity linking to iteratively retrieve and construct a graph of documents. These iterative retrieval methods mitigate the recall problem of one-shot retrieval, however, without iterative reranking and filtering process, the expanded documents inevitably introduce noise to the downstream QA model. Multi-step Reasoner~\cite{das2018multistep} and MUPPET~\cite{feldman-el-yaniv-2019-multi} read retrieved documents to reformulate the query in latent space for iterative retrieval. However, these embedding-based retrieval methods have difficulties to capture the lexical information in entities due to the compression of information into embedding space. Moreover, these methods perform a fixed number of retrieval steps, which are not able to handle questions that require arbitrary hops of reasoning. Recurrent Retriever~\cite{asai2020learning} supports adaptive retrieval steps, but can only select one document at each step and has no interactions among documents outside of the retrieval chain. In contrast, our method leverages document graph instead of chain to propagate information, reranks and filters documents at each hop of retrieval, and terminates the retrieval process according to the number of positive retrieved documents in the graph.

\BlankLine
\noindent \textbf{Graph Neural Networks for QA.} Encouraged by the success of convolutional neural networks (CNNs) and recurrent neural networks (RNNs), graph neural networks (GNNs) are proposed to generalize both of them and organize the connections of neurons according to the structure of the input data~\cite{battaglia2018relational,wu2020comprehensive}. The main idea is to generate a node's representation by aggregating its own features and neighbors' features. Similar to CNNs, in GNNs, multiple graph convolutional layers can be stacked to produce high-level node representations. Many popular variants of GNNs are proposed, such as GraphSage~\cite{hamilton2017inductive}, Graph Convolutional Network (GCN)~\cite{Kipf:2016tc}, Graph Attention Network (GAT)~\cite{velickovic2018graph}, etc. GNNs have achieved great success on graph-related tasks such as node classification, node regression and graph classification.
Recently, Graph neural networks (GNNs) have been shown effective on knowledge-based QA tasks by reasoning over graphs~\cite{de-cao-etal-2019-question,sorokin-gurevych-2018-modeling,zhang2018variational}. Recent studies on text-based QA also leverage GNNs for multi-hop reasoning. HDE-Graph~\cite{tu-etal-2019-multi} constructs the graph with entity and document nodes to enable rich information interaction. CogQA~\cite{ding-etal-2019-cognitive} extracts candidate answer spans and next-hop entities to build the cognitive graph for reasoning. HGN~\cite{HGN2019} employs a hierarchical graph that consists of paragraph, sentence and entity nodes for reasoning on different granularities. DFGN~\cite{qiu-etal-2019-dynamically} introduces a fusion layer on top of the entity graph with a mask prediction module. Graph Retriever~\cite{min2019knowledge} uses graph convolution network to fuse information on the entity-linked graph of passages. Multi-grained MRC~\cite{zheng-etal-2020-document} utilizes GATs to obtain different levels of representations of documents for machine reading comprehension. These GNN-based methods serve as the \textit{reader} module to extract answers from a few documents. In contrast, our work employs graph attention network, a popular GNN variant, in the \textit{retriever} module. To the best of our knowledge, we are the first to propose GNN-based document reranking method for open-domain multi-hop QA.

\section{Conclusion}

We present a QA framework that can answer any-hop open-domain questions, which iteratively retrieves, reranks and filters documents with a graph-based reranking model, and adaptively decides how many steps of retrieval and reranking are needed for a multi-hop question. Our method consistently achieves promising performance on both single- and multi-hop open-domain QA datasets.

\tiny
\bibliographystyle{ACM-Reference-Format}
\bibliography{ref}


\begin{thebibliography}{44}


\ifx \showCODEN    \undefined \def \showCODEN     #1{\unskip}     \fi
\ifx \showDOI      \undefined \def \showDOI       #1{#1}\fi
\ifx \showISBNx    \undefined \def \showISBNx     #1{\unskip}     \fi
\ifx \showISBNxiii \undefined \def \showISBNxiii  #1{\unskip}     \fi
\ifx \showISSN     \undefined \def \showISSN      #1{\unskip}     \fi
\ifx \showLCCN     \undefined \def \showLCCN      #1{\unskip}     \fi
\ifx \shownote     \undefined \def \shownote      #1{#1}          \fi
\ifx \showarticletitle \undefined \def \showarticletitle #1{#1}   \fi
\ifx \showURL      \undefined \def \showURL       {\relax}        \fi
\providecommand\bibfield[2]{#2}
\providecommand\bibinfo[2]{#2}
\providecommand\natexlab[1]{#1}
\providecommand\showeprint[2][]{arXiv:#2}

\bibitem[\protect\citeauthoryear{Asai, Hashimoto, Hajishirzi, Socher, and
  Xiong}{Asai et~al\mbox{.}}{2020}]%
        {asai2020learning}
\bibfield{author}{\bibinfo{person}{Akari Asai}, \bibinfo{person}{Kazuma
  Hashimoto}, \bibinfo{person}{Hannaneh Hajishirzi}, \bibinfo{person}{Richard
  Socher}, {and} \bibinfo{person}{Caiming Xiong}.}
  \bibinfo{year}{2020}\natexlab{}.
\newblock \showarticletitle{Learning to Retrieve Reasoning Paths over Wikipedia
  Graph for Question Answering}. In \bibinfo{booktitle}{\emph{International
  Conference on Learning Representations}}.
\newblock


\bibitem[\protect\citeauthoryear{Battaglia, Hamrick, Bapst, Sanchez-Gonzalez,
  Zambaldi, Malinowski, Tacchetti, Raposo, Santoro, Faulkner,
  et~al\mbox{.}}{Battaglia et~al\mbox{.}}{2018}]%
        {battaglia2018relational}
\bibfield{author}{\bibinfo{person}{Peter~W Battaglia},
  \bibinfo{person}{Jessica~B Hamrick}, \bibinfo{person}{Victor Bapst},
  \bibinfo{person}{Alvaro Sanchez-Gonzalez}, \bibinfo{person}{Vinicius
  Zambaldi}, \bibinfo{person}{Mateusz Malinowski}, \bibinfo{person}{Andrea
  Tacchetti}, \bibinfo{person}{David Raposo}, \bibinfo{person}{Adam Santoro},
  \bibinfo{person}{Ryan Faulkner}, {et~al\mbox{.}}}
  \bibinfo{year}{2018}\natexlab{}.
\newblock \showarticletitle{Relational inductive biases, deep learning, and
  graph networks}.
\newblock \bibinfo{journal}{\emph{arXiv}} (\bibinfo{year}{2018}).
\newblock


\bibitem[\protect\citeauthoryear{Chen, Fisch, Weston, and Bordes}{Chen
  et~al\mbox{.}}{2017}]%
        {chen-etal-2017-reading}
\bibfield{author}{\bibinfo{person}{Danqi Chen}, \bibinfo{person}{Adam Fisch},
  \bibinfo{person}{Jason Weston}, {and} \bibinfo{person}{Antoine Bordes}.}
  \bibinfo{year}{2017}\natexlab{}.
\newblock \showarticletitle{Reading {W}ikipedia to Answer Open-Domain
  Questions}. In \bibinfo{booktitle}{\emph{Proceedings of the 55th Annual
  Meeting of the Association for Computational Linguistics (Volume 1: Long
  Papers)}}. \bibinfo{address}{Vancouver, Canada}, \bibinfo{pages}{1870--1879}.
\newblock


\bibitem[\protect\citeauthoryear{Das, Dhuliawala, Zaheer, and McCallum}{Das
  et~al\mbox{.}}{2019a}]%
        {das2018multistep}
\bibfield{author}{\bibinfo{person}{Rajarshi Das}, \bibinfo{person}{Shehzaad
  Dhuliawala}, \bibinfo{person}{Manzil Zaheer}, {and} \bibinfo{person}{Andrew
  McCallum}.} \bibinfo{year}{2019}\natexlab{a}.
\newblock \showarticletitle{Multi-step Retriever-Reader Interaction for
  Scalable Open-domain Question Answering}. In
  \bibinfo{booktitle}{\emph{International Conference on Learning
  Representations}}.
\newblock


\bibitem[\protect\citeauthoryear{Das, Godbole, Kavarthapu, Gong, Singhal, Yu,
  Guo, Gao, Zamani, Zaheer, and McCallum}{Das et~al\mbox{.}}{2019b}]%
        {das-etal-2019-multi}
\bibfield{author}{\bibinfo{person}{Rajarshi Das}, \bibinfo{person}{Ameya
  Godbole}, \bibinfo{person}{Dilip Kavarthapu}, \bibinfo{person}{Zhiyu Gong},
  \bibinfo{person}{Abhishek Singhal}, \bibinfo{person}{Mo Yu},
  \bibinfo{person}{Xiaoxiao Guo}, \bibinfo{person}{Tian Gao},
  \bibinfo{person}{Hamed Zamani}, \bibinfo{person}{Manzil Zaheer}, {and}
  \bibinfo{person}{Andrew McCallum}.} \bibinfo{year}{2019}\natexlab{b}.
\newblock \showarticletitle{Multi-step Entity-centric Information Retrieval for
  Multi-Hop Question Answering}. In \bibinfo{booktitle}{\emph{Proceedings of
  the 2nd Workshop on Machine Reading for Question Answering}}.
  \bibinfo{address}{Hong Kong, China}, \bibinfo{pages}{113--118}.
\newblock


\bibitem[\protect\citeauthoryear{De~Cao, Aziz, and Titov}{De~Cao
  et~al\mbox{.}}{2019}]%
        {de-cao-etal-2019-question}
\bibfield{author}{\bibinfo{person}{Nicola De~Cao}, \bibinfo{person}{Wilker
  Aziz}, {and} \bibinfo{person}{Ivan Titov}.} \bibinfo{year}{2019}\natexlab{}.
\newblock \showarticletitle{Question Answering by Reasoning Across Documents
  with Graph Convolutional Networks}. In \bibinfo{booktitle}{\emph{Proceedings
  of the 2019 Conference of the North {A}merican Chapter of the Association for
  Computational Linguistics: Human Language Technologies, Volume 1 (Long and
  Short Papers)}}. \bibinfo{address}{Minneapolis, Minnesota},
  \bibinfo{pages}{2306--2317}.
\newblock


\bibitem[\protect\citeauthoryear{Devlin, Chang, Lee, and Toutanova}{Devlin
  et~al\mbox{.}}{2019}]%
        {devlin-etal-2019-bert}
\bibfield{author}{\bibinfo{person}{Jacob Devlin}, \bibinfo{person}{Ming-Wei
  Chang}, \bibinfo{person}{Kenton Lee}, {and} \bibinfo{person}{Kristina
  Toutanova}.} \bibinfo{year}{2019}\natexlab{}.
\newblock \showarticletitle{{BERT}: Pre-training of Deep Bidirectional
  Transformers for Language Understanding}. In
  \bibinfo{booktitle}{\emph{Proceedings of the 2019 Conference of the North
  {A}merican Chapter of the Association for Computational Linguistics: Human
  Language Technologies, Volume 1 (Long and Short Papers)}}.
  \bibinfo{address}{Minneapolis, Minnesota}, \bibinfo{pages}{4171--4186}.
\newblock


\bibitem[\protect\citeauthoryear{Dhingra, Zaheer, Balachandran, Neubig,
  Salakhutdinov, and Cohen}{Dhingra et~al\mbox{.}}{2020}]%
        {Dhingra2020Differentiable}
\bibfield{author}{\bibinfo{person}{Bhuwan Dhingra}, \bibinfo{person}{Manzil
  Zaheer}, \bibinfo{person}{Vidhisha Balachandran}, \bibinfo{person}{Graham
  Neubig}, \bibinfo{person}{Ruslan Salakhutdinov}, {and}
  \bibinfo{person}{William~W. Cohen}.} \bibinfo{year}{2020}\natexlab{}.
\newblock \showarticletitle{Differentiable Reasoning over a Virtual Knowledge
  Base}. In \bibinfo{booktitle}{\emph{International Conference on Learning
  Representations}}.
\newblock


\bibitem[\protect\citeauthoryear{Ding, Zhou, Chen, Yang, and Tang}{Ding
  et~al\mbox{.}}{2019}]%
        {ding-etal-2019-cognitive}
\bibfield{author}{\bibinfo{person}{Ming Ding}, \bibinfo{person}{Chang Zhou},
  \bibinfo{person}{Qibin Chen}, \bibinfo{person}{Hongxia Yang}, {and}
  \bibinfo{person}{Jie Tang}.} \bibinfo{year}{2019}\natexlab{}.
\newblock \showarticletitle{Cognitive Graph for Multi-Hop Reading Comprehension
  at Scale}. In \bibinfo{booktitle}{\emph{Proceedings of the 57th Annual
  Meeting of the Association for Computational Linguistics}}.
  \bibinfo{publisher}{Association for Computational Linguistics},
  \bibinfo{address}{Florence, Italy}, \bibinfo{pages}{2694--2703}.
\newblock


\bibitem[\protect\citeauthoryear{Fang, Sun, Gan, Pillai, Wang, and Liu}{Fang
  et~al\mbox{.}}{2020}]%
        {Fang2019HierarchicalGN}
\bibfield{author}{\bibinfo{person}{Yuwei Fang}, \bibinfo{person}{Siqi Sun},
  \bibinfo{person}{Zhe Gan}, \bibinfo{person}{Rohit Pillai},
  \bibinfo{person}{Shuohang Wang}, {and} \bibinfo{person}{Jingjing Liu}.}
  \bibinfo{year}{2020}\natexlab{}.
\newblock \showarticletitle{Hierarchical Graph Network for Multi-hop Question
  Answering}. In \bibinfo{booktitle}{\emph{Proceedings of the 2020 Conference
  on Empirical Methods in Natural Language Processing (EMNLP)}}.
  \bibinfo{address}{Online}, \bibinfo{pages}{8823--8838}.
\newblock


\bibitem[\protect\citeauthoryear{Feldman and El-Yaniv}{Feldman and
  El-Yaniv}{2019}]%
        {feldman-el-yaniv-2019-multi}
\bibfield{author}{\bibinfo{person}{Yair Feldman} {and} \bibinfo{person}{Ran
  El-Yaniv}.} \bibinfo{year}{2019}\natexlab{}.
\newblock \showarticletitle{Multi-Hop Paragraph Retrieval for Open-Domain
  Question Answering}. In \bibinfo{booktitle}{\emph{Proceedings of the 57th
  Annual Meeting of the Association for Computational Linguistics}}.
  \bibinfo{address}{Florence, Italy}, \bibinfo{pages}{2296--2309}.
\newblock


\bibitem[\protect\citeauthoryear{Ferrucci, Brown, Chu-Carroll, Fan, Gondek,
  Kalyanpur, Lally, Murdock, Nyberg, Prager, et~al\mbox{.}}{Ferrucci
  et~al\mbox{.}}{2010}]%
        {ferrucci2010building}
\bibfield{author}{\bibinfo{person}{David Ferrucci}, \bibinfo{person}{Eric
  Brown}, \bibinfo{person}{Jennifer Chu-Carroll}, \bibinfo{person}{James Fan},
  \bibinfo{person}{David Gondek}, \bibinfo{person}{Aditya~A Kalyanpur},
  \bibinfo{person}{Adam Lally}, \bibinfo{person}{J~William Murdock},
  \bibinfo{person}{Eric Nyberg}, \bibinfo{person}{John Prager},
  {et~al\mbox{.}}} \bibinfo{year}{2010}\natexlab{}.
\newblock \showarticletitle{Building Watson: An overview of the DeepQA
  project}.
\newblock \bibinfo{journal}{\emph{AI magazine}} \bibinfo{volume}{31},
  \bibinfo{number}{3} (\bibinfo{year}{2010}), \bibinfo{pages}{59--79}.
\newblock


\bibitem[\protect\citeauthoryear{Hamilton, Ying, and Leskovec}{Hamilton
  et~al\mbox{.}}{2017}]%
        {hamilton2017inductive}
\bibfield{author}{\bibinfo{person}{Will Hamilton}, \bibinfo{person}{Zhitao
  Ying}, {and} \bibinfo{person}{Jure Leskovec}.}
  \bibinfo{year}{2017}\natexlab{}.
\newblock \showarticletitle{Inductive representation learning on large graphs}.
  In \bibinfo{booktitle}{\emph{Advances in neural information processing
  systems}}. \bibinfo{pages}{1024--1034}.
\newblock


\bibitem[\protect\citeauthoryear{Karpukhin, O{\u{g}}uz, Min, Wu, Edunov, Chen,
  and Yih}{Karpukhin et~al\mbox{.}}{2020}]%
        {karpukhin2020dense}
\bibfield{author}{\bibinfo{person}{Vladimir Karpukhin}, \bibinfo{person}{Barlas
  O{\u{g}}uz}, \bibinfo{person}{Sewon Min}, \bibinfo{person}{Ledell Wu},
  \bibinfo{person}{Sergey Edunov}, \bibinfo{person}{Danqi Chen}, {and}
  \bibinfo{person}{Wen-tau Yih}.} \bibinfo{year}{2020}\natexlab{}.
\newblock \showarticletitle{Dense Passage Retrieval for Open-Domain Question
  Answering}.
\newblock \bibinfo{journal}{\emph{arXiv preprint arXiv:2004.04906}}
  (\bibinfo{year}{2020}).
\newblock


\bibitem[\protect\citeauthoryear{Kipf and Welling}{Kipf and Welling}{2017}]%
        {Kipf:2016tc}
\bibfield{author}{\bibinfo{person}{Thomas~N. Kipf} {and} \bibinfo{person}{Max
  Welling}.} \bibinfo{year}{2017}\natexlab{}.
\newblock \showarticletitle{{Semi-Supervised Classification with Graph
  Convolutional Networks}}. In \bibinfo{booktitle}{\emph{Proceedings of the 5th
  International Conference on Learning Representations}} (Palais des
  Congr{\`e}s Neptune, Toulon, France) \emph{(\bibinfo{series}{ICLR '17})}.
\newblock
\urldef\tempurl%
\url{https://openreview.net/forum?id=SJU4ayYgl}
\showURL{%
\tempurl}


\bibitem[\protect\citeauthoryear{Lan, Chen, Goodman, Gimpel, Sharma, and
  Soricut}{Lan et~al\mbox{.}}{2020}]%
        {Lan2020ALBERT:}
\bibfield{author}{\bibinfo{person}{Zhenzhong Lan}, \bibinfo{person}{Mingda
  Chen}, \bibinfo{person}{Sebastian Goodman}, \bibinfo{person}{Kevin Gimpel},
  \bibinfo{person}{Piyush Sharma}, {and} \bibinfo{person}{Radu Soricut}.}
  \bibinfo{year}{2020}\natexlab{}.
\newblock \showarticletitle{ALBERT: A Lite BERT for Self-supervised Learning of
  Language Representations}. In \bibinfo{booktitle}{\emph{International
  Conference on Learning Representations}}.
\newblock


\bibitem[\protect\citeauthoryear{Lee, Chang, and Toutanova}{Lee
  et~al\mbox{.}}{2019}]%
        {lee-etal-2019-latent}
\bibfield{author}{\bibinfo{person}{Kenton Lee}, \bibinfo{person}{Ming-Wei
  Chang}, {and} \bibinfo{person}{Kristina Toutanova}.}
  \bibinfo{year}{2019}\natexlab{}.
\newblock \showarticletitle{Latent Retrieval for Weakly Supervised Open Domain
  Question Answering}. In \bibinfo{booktitle}{\emph{Proceedings of the 57th
  Annual Meeting of the Association for Computational Linguistics}}.
  \bibinfo{publisher}{Association for Computational Linguistics},
  \bibinfo{address}{Florence, Italy}, \bibinfo{pages}{6086--6096}.
\newblock


\bibitem[\protect\citeauthoryear{Li, Jia, Shen, Shi, and Yang}{Li
  et~al\mbox{.}}{2019}]%
        {HGN2019}
\bibfield{author}{\bibinfo{person}{Chong Li}, \bibinfo{person}{Kunyang Jia},
  \bibinfo{person}{Dan Shen}, \bibinfo{person}{C.J.~Richard Shi}, {and}
  \bibinfo{person}{Hongxia Yang}.} \bibinfo{year}{2019}\natexlab{}.
\newblock \showarticletitle{Hierarchical Representation Learning for Bipartite
  Graphs}. In \bibinfo{booktitle}{\emph{Proceedings of the Twenty-Eighth
  International Joint Conference on Artificial Intelligence, {IJCAI-19}}}.
  \bibinfo{publisher}{International Joint Conferences on Artificial
  Intelligence Organization}.
\newblock


\bibitem[\protect\citeauthoryear{Li, Li, Shang, Jiang, Liu, Sun, Ji, and
  Liu}{Li et~al\mbox{.}}{2020}]%
        {li2020hopretriever}
\bibfield{author}{\bibinfo{person}{Shaobo Li}, \bibinfo{person}{Xiaoguang Li},
  \bibinfo{person}{Lifeng Shang}, \bibinfo{person}{Xin Jiang},
  \bibinfo{person}{Qun Liu}, \bibinfo{person}{Chengjie Sun},
  \bibinfo{person}{Zhenzhou Ji}, {and} \bibinfo{person}{Bingquan Liu}.}
  \bibinfo{year}{2020}\natexlab{}.
\newblock \showarticletitle{HopRetriever: Retrieve Hops over Wikipedia to
  Answer Complex Questions}.
\newblock \bibinfo{journal}{\emph{arXiv preprint arXiv:2012.15534}}
  (\bibinfo{year}{2020}).
\newblock


\bibitem[\protect\citeauthoryear{Min, Chen, Zettlemoyer, and Hajishirzi}{Min
  et~al\mbox{.}}{2019a}]%
        {min2019knowledge}
\bibfield{author}{\bibinfo{person}{Sewon Min}, \bibinfo{person}{Danqi Chen},
  \bibinfo{person}{Luke Zettlemoyer}, {and} \bibinfo{person}{Hannaneh
  Hajishirzi}.} \bibinfo{year}{2019}\natexlab{a}.
\newblock \showarticletitle{Knowledge guided text retrieval and reading for
  open domain QA}.
\newblock \bibinfo{journal}{\emph{arXiv preprint arXiv:1911.03868}}
  (\bibinfo{year}{2019}).
\newblock


\bibitem[\protect\citeauthoryear{Min, Zhong, Zettlemoyer, and Hajishirzi}{Min
  et~al\mbox{.}}{2019b}]%
        {min-etal-2019-multi}
\bibfield{author}{\bibinfo{person}{Sewon Min}, \bibinfo{person}{Victor Zhong},
  \bibinfo{person}{Luke Zettlemoyer}, {and} \bibinfo{person}{Hannaneh
  Hajishirzi}.} \bibinfo{year}{2019}\natexlab{b}.
\newblock \showarticletitle{Multi-hop Reading Comprehension through Question
  Decomposition and Rescoring}. In \bibinfo{booktitle}{\emph{Proceedings of the
  57th Annual Meeting of the Association for Computational Linguistics}}.
  \bibinfo{publisher}{Association for Computational Linguistics},
  \bibinfo{address}{Florence, Italy}, \bibinfo{pages}{6097--6109}.
\newblock


\bibitem[\protect\citeauthoryear{Ni, Zhu, Chen, and McAuley}{Ni
  et~al\mbox{.}}{2019}]%
        {ni-etal-2019-learning}
\bibfield{author}{\bibinfo{person}{Jianmo Ni}, \bibinfo{person}{Chenguang Zhu},
  \bibinfo{person}{Weizhu Chen}, {and} \bibinfo{person}{Julian McAuley}.}
  \bibinfo{year}{2019}\natexlab{}.
\newblock \showarticletitle{Learning to Attend On Essential Terms: An Enhanced
  Retriever-Reader Model for Open-domain Question Answering}. In
  \bibinfo{booktitle}{\emph{Proceedings of the 2019 Conference of the North
  {A}merican Chapter of the Association for Computational Linguistics: Human
  Language Technologies, Volume 1 (Long and Short Papers)}}.
  \bibinfo{publisher}{Association for Computational Linguistics},
  \bibinfo{address}{Minneapolis, Minnesota}, \bibinfo{pages}{335--344}.
\newblock


\bibitem[\protect\citeauthoryear{Nie, Wang, and Bansal}{Nie
  et~al\mbox{.}}{2019}]%
        {nie-etal-2019-revealing}
\bibfield{author}{\bibinfo{person}{Yixin Nie}, \bibinfo{person}{Songhe Wang},
  {and} \bibinfo{person}{Mohit Bansal}.} \bibinfo{year}{2019}\natexlab{}.
\newblock \showarticletitle{Revealing the Importance of Semantic Retrieval for
  Machine Reading at Scale}. In \bibinfo{booktitle}{\emph{Proceedings of the
  2019 Conference on Empirical Methods in Natural Language Processing and the
  9th International Joint Conference on Natural Language Processing
  (EMNLP-IJCNLP)}}. \bibinfo{publisher}{Association for Computational
  Linguistics}, \bibinfo{address}{Hong Kong, China},
  \bibinfo{pages}{2553--2566}.
\newblock


\bibitem[\protect\citeauthoryear{Nishida, Nishida, Nagata, Otsuka, Saito,
  Asano, and Tomita}{Nishida et~al\mbox{.}}{2019}]%
        {nishida-etal-2019-answering}
\bibfield{author}{\bibinfo{person}{Kosuke Nishida}, \bibinfo{person}{Kyosuke
  Nishida}, \bibinfo{person}{Masaaki Nagata}, \bibinfo{person}{Atsushi Otsuka},
  \bibinfo{person}{Itsumi Saito}, \bibinfo{person}{Hisako Asano}, {and}
  \bibinfo{person}{Junji Tomita}.} \bibinfo{year}{2019}\natexlab{}.
\newblock \showarticletitle{Answering while Summarizing: Multi-task Learning
  for Multi-hop {QA} with Evidence Extraction}. In
  \bibinfo{booktitle}{\emph{Proceedings of the 57th Annual Meeting of the
  Association for Computational Linguistics}}. \bibinfo{publisher}{Association
  for Computational Linguistics}, \bibinfo{address}{Florence, Italy},
  \bibinfo{pages}{2335--2345}.
\newblock


\bibitem[\protect\citeauthoryear{Qi, Lin, Mehr, Wang, and Manning}{Qi
  et~al\mbox{.}}{2019}]%
        {qi-etal-2019-answering}
\bibfield{author}{\bibinfo{person}{Peng Qi}, \bibinfo{person}{Xiaowen Lin},
  \bibinfo{person}{Leo Mehr}, \bibinfo{person}{Zijian Wang}, {and}
  \bibinfo{person}{Christopher~D. Manning}.} \bibinfo{year}{2019}\natexlab{}.
\newblock \showarticletitle{Answering Complex Open-domain Questions Through
  Iterative Query Generation}. In \bibinfo{booktitle}{\emph{Proceedings of the
  2019 Conference on Empirical Methods in Natural Language Processing and the
  9th International Joint Conference on Natural Language Processing
  (EMNLP-IJCNLP)}}. \bibinfo{publisher}{Association for Computational
  Linguistics}, \bibinfo{address}{Hong Kong, China},
  \bibinfo{pages}{2590--2602}.
\newblock


\bibitem[\protect\citeauthoryear{Qiu, Xiao, Qu, Zhou, Li, Zhang, and Yu}{Qiu
  et~al\mbox{.}}{2019}]%
        {qiu-etal-2019-dynamically}
\bibfield{author}{\bibinfo{person}{Lin Qiu}, \bibinfo{person}{Yunxuan Xiao},
  \bibinfo{person}{Yanru Qu}, \bibinfo{person}{Hao Zhou}, \bibinfo{person}{Lei
  Li}, \bibinfo{person}{Weinan Zhang}, {and} \bibinfo{person}{Yong Yu}.}
  \bibinfo{year}{2019}\natexlab{}.
\newblock \showarticletitle{Dynamically Fused Graph Network for Multi-hop
  Reasoning}. In \bibinfo{booktitle}{\emph{Proceedings of the 57th Annual
  Meeting of the Association for Computational Linguistics}}.
  \bibinfo{address}{Florence, Italy}, \bibinfo{pages}{6140--6150}.
\newblock


\bibitem[\protect\citeauthoryear{Rajpurkar, Zhang, Lopyrev, and
  Liang}{Rajpurkar et~al\mbox{.}}{2016}]%
        {rajpurkar-etal-2016-squad}
\bibfield{author}{\bibinfo{person}{Pranav Rajpurkar}, \bibinfo{person}{Jian
  Zhang}, \bibinfo{person}{Konstantin Lopyrev}, {and} \bibinfo{person}{Percy
  Liang}.} \bibinfo{year}{2016}\natexlab{}.
\newblock \showarticletitle{{SQ}u{AD}: 100,000+ Questions for Machine
  Comprehension of Text}. In \bibinfo{booktitle}{\emph{Proceedings of the 2016
  Conference on Empirical Methods in Natural Language Processing}}.
  \bibinfo{publisher}{Association for Computational Linguistics},
  \bibinfo{address}{Austin, Texas}, \bibinfo{pages}{2383--2392}.
\newblock


\bibitem[\protect\citeauthoryear{Sorokin and Gurevych}{Sorokin and
  Gurevych}{2018}]%
        {sorokin-gurevych-2018-modeling}
\bibfield{author}{\bibinfo{person}{Daniil Sorokin} {and} \bibinfo{person}{Iryna
  Gurevych}.} \bibinfo{year}{2018}\natexlab{}.
\newblock \showarticletitle{Modeling Semantics with Gated Graph Neural Networks
  for Knowledge Base Question Answering}. In
  \bibinfo{booktitle}{\emph{Proceedings of the 27th International Conference on
  Computational Linguistics}}. \bibinfo{publisher}{Association for
  Computational Linguistics}, \bibinfo{address}{Santa Fe, New Mexico, USA},
  \bibinfo{pages}{3306--3317}.
\newblock


\bibitem[\protect\citeauthoryear{Tu, Wang, Huang, Tang, He, and Zhou}{Tu
  et~al\mbox{.}}{2019}]%
        {tu-etal-2019-multi}
\bibfield{author}{\bibinfo{person}{Ming Tu}, \bibinfo{person}{Guangtao Wang},
  \bibinfo{person}{Jing Huang}, \bibinfo{person}{Yun Tang},
  \bibinfo{person}{Xiaodong He}, {and} \bibinfo{person}{Bowen Zhou}.}
  \bibinfo{year}{2019}\natexlab{}.
\newblock \showarticletitle{Multi-hop Reading Comprehension across Multiple
  Documents by Reasoning over Heterogeneous Graphs}. In
  \bibinfo{booktitle}{\emph{Proceedings of the 57th Annual Meeting of the
  Association for Computational Linguistics}}. \bibinfo{publisher}{Association
  for Computational Linguistics}, \bibinfo{address}{Florence, Italy},
  \bibinfo{pages}{2704--2713}.
\newblock


\bibitem[\protect\citeauthoryear{Vaswani, Shazeer, Parmar, Uszkoreit, Jones,
  Gomez, Kaiser, and Polosukhin}{Vaswani et~al\mbox{.}}{2017}]%
        {vaswani2017attention}
\bibfield{author}{\bibinfo{person}{Ashish Vaswani}, \bibinfo{person}{Noam
  Shazeer}, \bibinfo{person}{Niki Parmar}, \bibinfo{person}{Jakob Uszkoreit},
  \bibinfo{person}{Llion Jones}, \bibinfo{person}{Aidan~N Gomez},
  \bibinfo{person}{\L~ukasz Kaiser}, {and} \bibinfo{person}{Illia Polosukhin}.}
  \bibinfo{year}{2017}\natexlab{}.
\newblock \showarticletitle{Attention is All you Need}.
\newblock In \bibinfo{booktitle}{\emph{Advances in Neural Information
  Processing Systems 30}}, \bibfield{editor}{\bibinfo{person}{I.~Guyon},
  \bibinfo{person}{U.~V. Luxburg}, \bibinfo{person}{S.~Bengio},
  \bibinfo{person}{H.~Wallach}, \bibinfo{person}{R.~Fergus},
  \bibinfo{person}{S.~Vishwanathan}, {and} \bibinfo{person}{R.~Garnett}}
  (Eds.). \bibinfo{publisher}{Curran Associates, Inc.},
  \bibinfo{pages}{5998--6008}.
\newblock


\bibitem[\protect\citeauthoryear{Veli{\v{c}}kovi{\'{c}}, Cucurull, Casanova,
  Romero, Li{\`{o}}, and Bengio}{Veli{\v{c}}kovi{\'{c}} et~al\mbox{.}}{2018}]%
        {velickovic2018graph}
\bibfield{author}{\bibinfo{person}{Petar Veli{\v{c}}kovi{\'{c}}},
  \bibinfo{person}{Guillem Cucurull}, \bibinfo{person}{Arantxa Casanova},
  \bibinfo{person}{Adriana Romero}, \bibinfo{person}{Pietro Li{\`{o}}}, {and}
  \bibinfo{person}{Yoshua Bengio}.} \bibinfo{year}{2018}\natexlab{}.
\newblock \showarticletitle{{Graph Attention Networks}}.
\newblock \bibinfo{journal}{\emph{International Conference on Learning
  Representations}} (\bibinfo{year}{2018}).
\newblock


\bibitem[\protect\citeauthoryear{Voorhees and Tice}{Voorhees and Tice}{2000}]%
        {voorhees2000building}
\bibfield{author}{\bibinfo{person}{Ellen~M Voorhees} {and}
  \bibinfo{person}{Dawn~M Tice}.} \bibinfo{year}{2000}\natexlab{}.
\newblock \showarticletitle{Building a question answering test collection}. In
  \bibinfo{booktitle}{\emph{Proceedings of the 23rd annual international ACM
  SIGIR conference on Research and development in information retrieval}}.
  \bibinfo{pages}{200--207}.
\newblock


\bibitem[\protect\citeauthoryear{Wang, Yao, Zhang, Xu, Tian, Liu, and
  Zhao}{Wang et~al\mbox{.}}{2019b}]%
        {wang2019sigir}
\bibfield{author}{\bibinfo{person}{Bingning Wang}, \bibinfo{person}{Ting Yao},
  \bibinfo{person}{Qi Zhang}, \bibinfo{person}{Jingfang Xu},
  \bibinfo{person}{Zhixing Tian}, \bibinfo{person}{Kang Liu}, {and}
  \bibinfo{person}{Jun Zhao}.} \bibinfo{year}{2019}\natexlab{b}.
\newblock \showarticletitle{Document Gated Reader for Open-Domain Question
  Answering}. In \bibinfo{booktitle}{\emph{Proceedings of the 42nd
  International ACM SIGIR Conference on Research and Development in Information
  Retrieval}} (Paris, France) \emph{(\bibinfo{series}{SIGIR’19})}.
  \bibinfo{publisher}{Association for Computing Machinery},
  \bibinfo{address}{New York, NY, USA}, \bibinfo{pages}{85–94}.
\newblock


\bibitem[\protect\citeauthoryear{Wang, Yu, Guo, Wang, Klinger, Zhang, Chang,
  Tesauro, Zhou, and Jiang}{Wang et~al\mbox{.}}{2018a}]%
        {wang2018r}
\bibfield{author}{\bibinfo{person}{Shuohang Wang}, \bibinfo{person}{Mo Yu},
  \bibinfo{person}{Xiaoxiao Guo}, \bibinfo{person}{Zhiguo Wang},
  \bibinfo{person}{Tim Klinger}, \bibinfo{person}{Wei Zhang},
  \bibinfo{person}{Shiyu Chang}, \bibinfo{person}{Gerry Tesauro},
  \bibinfo{person}{Bowen Zhou}, {and} \bibinfo{person}{Jing Jiang}.}
  \bibinfo{year}{2018}\natexlab{a}.
\newblock \showarticletitle{R$^3$: Reinforced Ranker-Reader for Open-Domain
  Question Answering}. In \bibinfo{booktitle}{\emph{Thirty-Second AAAI
  Conference on Artificial Intelligence}}.
\newblock


\bibitem[\protect\citeauthoryear{Wang, Yu, Jiang, Zhang, Guo, Chang, Wang,
  Klinger, Tesauro, and Campbell}{Wang et~al\mbox{.}}{2018b}]%
        {wang2018evidence}
\bibfield{author}{\bibinfo{person}{Shuohang Wang}, \bibinfo{person}{Mo Yu},
  \bibinfo{person}{Jing Jiang}, \bibinfo{person}{Wei Zhang},
  \bibinfo{person}{Xiaoxiao Guo}, \bibinfo{person}{Shiyu Chang},
  \bibinfo{person}{Zhiguo Wang}, \bibinfo{person}{Tim Klinger},
  \bibinfo{person}{Gerald Tesauro}, {and} \bibinfo{person}{Murray Campbell}.}
  \bibinfo{year}{2018}\natexlab{b}.
\newblock \showarticletitle{Evidence Aggregation for Answer Re-Ranking in
  Open-Domain Question Answering}. In \bibinfo{booktitle}{\emph{International
  Conference on Learning Representations}}.
\newblock


\bibitem[\protect\citeauthoryear{Wang, Ng, Ma, Nallapati, and Xiang}{Wang
  et~al\mbox{.}}{2019a}]%
        {wang-etal-2019-multi}
\bibfield{author}{\bibinfo{person}{Zhiguo Wang}, \bibinfo{person}{Patrick Ng},
  \bibinfo{person}{Xiaofei Ma}, \bibinfo{person}{Ramesh Nallapati}, {and}
  \bibinfo{person}{Bing Xiang}.} \bibinfo{year}{2019}\natexlab{a}.
\newblock \showarticletitle{Multi-passage {BERT}: A Globally Normalized {BERT}
  Model for Open-domain Question Answering}. In
  \bibinfo{booktitle}{\emph{Proceedings of the 2019 Conference on Empirical
  Methods in Natural Language Processing and the 9th International Joint
  Conference on Natural Language Processing (EMNLP-IJCNLP)}}.
  \bibinfo{publisher}{Association for Computational Linguistics},
  \bibinfo{address}{Hong Kong, China}, \bibinfo{pages}{5878--5882}.
\newblock


\bibitem[\protect\citeauthoryear{Wolf, Debut, Sanh, Chaumond, Delangue, Moi,
  Cistac, Rault, Louf, Funtowicz, and Brew}{Wolf et~al\mbox{.}}{2019}]%
        {Wolf2019HuggingFacesTS}
\bibfield{author}{\bibinfo{person}{Thomas Wolf}, \bibinfo{person}{Lysandre
  Debut}, \bibinfo{person}{Victor Sanh}, \bibinfo{person}{Julien Chaumond},
  \bibinfo{person}{Clement Delangue}, \bibinfo{person}{Anthony Moi},
  \bibinfo{person}{Pierric Cistac}, \bibinfo{person}{Tim Rault},
  \bibinfo{person}{R'emi Louf}, \bibinfo{person}{Morgan Funtowicz}, {and}
  \bibinfo{person}{Jamie Brew}.} \bibinfo{year}{2019}\natexlab{}.
\newblock \showarticletitle{Hugging{F}ace's Transformers: State-of-the-art
  Natural Language Processing}.
\newblock \bibinfo{journal}{\emph{ArXiv}}  \bibinfo{volume}{abs/1910.03771}
  (\bibinfo{year}{2019}).
\newblock


\bibitem[\protect\citeauthoryear{Wu, Pan, Chen, Long, Zhang, and Philip}{Wu
  et~al\mbox{.}}{2020}]%
        {wu2020comprehensive}
\bibfield{author}{\bibinfo{person}{Zonghan Wu}, \bibinfo{person}{Shirui Pan},
  \bibinfo{person}{Fengwen Chen}, \bibinfo{person}{Guodong Long},
  \bibinfo{person}{Chengqi Zhang}, {and} \bibinfo{person}{S~Yu Philip}.}
  \bibinfo{year}{2020}\natexlab{}.
\newblock \showarticletitle{A comprehensive survey on graph neural networks}.
\newblock \bibinfo{journal}{\emph{IEEE Transactions on Neural Networks and
  Learning Systems}} (\bibinfo{year}{2020}).
\newblock


\bibitem[\protect\citeauthoryear{Xiong, Li, Iyer, Du, Lewis, Wang, Mehdad, Yih,
  Riedel, Kiela, and Oguz}{Xiong et~al\mbox{.}}{2021}]%
        {xiong2021answering}
\bibfield{author}{\bibinfo{person}{Wenhan Xiong}, \bibinfo{person}{Xiang Li},
  \bibinfo{person}{Srini Iyer}, \bibinfo{person}{Jingfei Du},
  \bibinfo{person}{Patrick Lewis}, \bibinfo{person}{William~Yang Wang},
  \bibinfo{person}{Yashar Mehdad}, \bibinfo{person}{Scott Yih},
  \bibinfo{person}{Sebastian Riedel}, \bibinfo{person}{Douwe Kiela}, {and}
  \bibinfo{person}{Barlas Oguz}.} \bibinfo{year}{2021}\natexlab{}.
\newblock \showarticletitle{Answering Complex Open-Domain Questions with
  Multi-Hop Dense Retrieval}. In \bibinfo{booktitle}{\emph{International
  Conference on Learning Representations}}.
\newblock
\urldef\tempurl%
\url{https://openreview.net/forum?id=EMHoBG0avc1}
\showURL{%
\tempurl}


\bibitem[\protect\citeauthoryear{Yang, Xie, Lin, Li, Tan, Xiong, Li, and
  Lin}{Yang et~al\mbox{.}}{2019}]%
        {yang-etal-2019-end-end-open}
\bibfield{author}{\bibinfo{person}{Wei Yang}, \bibinfo{person}{Yuqing Xie},
  \bibinfo{person}{Aileen Lin}, \bibinfo{person}{Xingyu Li},
  \bibinfo{person}{Luchen Tan}, \bibinfo{person}{Kun Xiong},
  \bibinfo{person}{Ming Li}, {and} \bibinfo{person}{Jimmy Lin}.}
  \bibinfo{year}{2019}\natexlab{}.
\newblock \showarticletitle{End-to-End Open-Domain Question Answering with
  {BERT}serini}. In \bibinfo{booktitle}{\emph{Proceedings of the 2019
  Conference of the North {A}merican Chapter of the Association for
  Computational Linguistics (Demonstrations)}}. \bibinfo{publisher}{Association
  for Computational Linguistics}, \bibinfo{address}{Minneapolis, Minnesota},
  \bibinfo{pages}{72--77}.
\newblock


\bibitem[\protect\citeauthoryear{Yang, Qi, Zhang, Bengio, Cohen, Salakhutdinov,
  and Manning}{Yang et~al\mbox{.}}{2018}]%
        {yang-etal-2018-hotpotqa}
\bibfield{author}{\bibinfo{person}{Zhilin Yang}, \bibinfo{person}{Peng Qi},
  \bibinfo{person}{Saizheng Zhang}, \bibinfo{person}{Yoshua Bengio},
  \bibinfo{person}{William Cohen}, \bibinfo{person}{Ruslan Salakhutdinov},
  {and} \bibinfo{person}{Christopher~D. Manning}.}
  \bibinfo{year}{2018}\natexlab{}.
\newblock \showarticletitle{{H}otpot{QA}: A Dataset for Diverse, Explainable
  Multi-hop Question Answering}. In \bibinfo{booktitle}{\emph{Proceedings of
  the 2018 Conference on Empirical Methods in Natural Language Processing}}.
  \bibinfo{publisher}{Association for Computational Linguistics},
  \bibinfo{address}{Brussels, Belgium}, \bibinfo{pages}{2369--2380}.
\newblock


\bibitem[\protect\citeauthoryear{Zhang, Dai, Kozareva, Smola, and Song}{Zhang
  et~al\mbox{.}}{2018}]%
        {zhang2018variational}
\bibfield{author}{\bibinfo{person}{Yuyu Zhang}, \bibinfo{person}{Hanjun Dai},
  \bibinfo{person}{Zornitsa Kozareva}, \bibinfo{person}{Alexander~J Smola},
  {and} \bibinfo{person}{Le Song}.} \bibinfo{year}{2018}\natexlab{}.
\newblock \showarticletitle{Variational reasoning for question answering with
  knowledge graph}. In \bibinfo{booktitle}{\emph{Thirty-Second AAAI Conference
  on Artificial Intelligence}}.
\newblock


\bibitem[\protect\citeauthoryear{Zhao, Xiong, Rosset, Song, Bennett, and
  Tiwary}{Zhao et~al\mbox{.}}{2020}]%
        {zhaotransxh2020}
\bibfield{author}{\bibinfo{person}{Chen Zhao}, \bibinfo{person}{Chenyan Xiong},
  \bibinfo{person}{Corby Rosset}, \bibinfo{person}{Xia Song},
  \bibinfo{person}{Paul Bennett}, {and} \bibinfo{person}{Saurabh Tiwary}.}
  \bibinfo{year}{2020}\natexlab{}.
\newblock \showarticletitle{Transformer-XH: Multi-Evidence Reasoning with eXtra
  Hop Attention}. In \bibinfo{booktitle}{\emph{International Conference on
  Learning Representations}}.
\newblock


\bibitem[\protect\citeauthoryear{Zheng, Wen, Liang, Duan, Che, Jiang, Zhou, and
  Liu}{Zheng et~al\mbox{.}}{2020}]%
        {zheng-etal-2020-document}
\bibfield{author}{\bibinfo{person}{Bo Zheng}, \bibinfo{person}{Haoyang Wen},
  \bibinfo{person}{Yaobo Liang}, \bibinfo{person}{Nan Duan},
  \bibinfo{person}{Wanxiang Che}, \bibinfo{person}{Daxin Jiang},
  \bibinfo{person}{Ming Zhou}, {and} \bibinfo{person}{Ting Liu}.}
  \bibinfo{year}{2020}\natexlab{}.
\newblock \showarticletitle{Document Modeling with Graph Attention Networks for
  Multi-grained Machine Reading Comprehension}. In
  \bibinfo{booktitle}{\emph{Proceedings of the 58th Annual Meeting of the
  Association for Computational Linguistics}}. \bibinfo{publisher}{Association
  for Computational Linguistics}, \bibinfo{address}{Online},
  \bibinfo{pages}{6708--6718}.
\newblock
\urldef\tempurl%
\url{https://doi.org/10.18653/v1/2020.acl-main.599}
\showDOI{\tempurl}


\end{thebibliography}

\end{document}